\pdfoutput=1
\documentclass[sigconf]{acmart}

\AtBeginDocument{%
  \providecommand\BibTeX{{%
    \normalfont B\kern-0.5em{\scshape i\kern-0.25em b}\kern-0.8em\TeX}}}

\setcopyright{acmcopyright}
\copyrightyear{2020}
\acmYear{2020}
\acmDOI{10.1145/1122445.1122456}

\acmConference[CIKM '20]{CIKM '20: International Conference on Information and Knowledge Management}{October 19--23, 2020}{Galway, Ireland}
\acmBooktitle{CIKM '20: ACM International Conference on Information and Knowledge Management,
  October 19--23, 2020, Galway, Ireland}
\acmPrice{15.00}
\acmISBN{978-1-4503-XXXX-X/18/06}

\usepackage{array,enumitem,multirow,caption,graphicx,subcaption,multicol,lipsum,float, tablefootnote}
\usepackage{amsmath, amsfonts, amsthm}
\usepackage{tabu,multirow}
\usepackage[ruled,vlined,linesnumbered]{algorithm2e}
\usepackage{kotex} 
\usepackage{color}
\usepackage{arydshln}
\usepackage{array,enumitem,multirow,caption,graphicx,subcaption,multicol,lipsum,float,adjustbox}
\usepackage{amsmath, amsfonts, amsthm}
\usepackage[colorinlistoftodos,textsize=tiny]{todonotes}
\newcolumntype{C}[1]{>{\centering\let\newline\\\arraybackslash\hspace{0pt}}m{#1}}
\newcolumntype{Z}{>{\centering\arraybackslash}m{0.062\linewidth}}
\newcommand{\RowStretch}[1]{\renewcommand{\arraystretch}{#1}}
\makeatletter
\def\thickhline{%
  \noalign{\ifnum0=`}\fi\hrule \@height \thickarrayrulewidth \futurelet
   \reserved@a\@xthickhline}
\def\@xthickhline{\ifx\reserved@a\thickhline
               \vskip\doublerulesep
               \vskip-\thickarrayrulewidth
             \fi
      \ifnum0=`{\fi}}
\makeatother
\newlength{\thickarrayrulewidth}
\copyrightyear{2020} 
\acmYear{2020} 
\setcopyright{acmcopyright}\acmConference[CIKM '20]{Proceedings of the 29th ACM International Conference on Information and Knowledge Management}{October 19--23, 2020}{Virtual Event, Ireland}
\acmBooktitle{Proceedings of the 29th ACM International Conference on Information and Knowledge Management (CIKM '20), October 19--23, 2020, Virtual Event, Ireland}
\acmPrice{15.00}
\acmDOI{10.1145/3340531.3412005}
\acmISBN{978-1-4503-6859-9/20/10}

\begin{document}
\setlength{\textfloatsep}{3pt}
\setlength{\intextsep}{3pt}
\title{DE-RRD: A Knowledge Distillation Framework \\ for Recommender System}
\fancyhead{}

\begin{abstract}
Recent recommender systems have started to employ knowledge distillation, which is a model compression technique distilling knowledge from a cumbersome model (teacher) to a compact model (student), to reduce inference latency while maintaining performance.
The state-of-the-art methods have only focused on making the student model accurately imitate the predictions of the teacher model.
They have a limitation in that the prediction results incompletely reveal the teacher’s knowledge.
In this paper, we propose a novel knowledge distillation framework for recommender system, called \textit{DE-RRD}, which enables the student model to learn from the latent knowledge encoded in the teacher model as well as from the teacher's predictions.
Concretely, \textit{DE-RRD} consists of two methods:
1) \textit{Distillation Experts (DE)} that directly transfers the latent knowledge from the teacher model.
DE exploits ``experts'' and a novel expert selection strategy for effectively distilling the vast teacher's knowledge to the student with limited capacity.
2) \textit{Relaxed Ranking Distillation (RRD)}
that transfers the knowledge revealed from the teacher's prediction with consideration of the relaxed ranking orders among items.
Our extensive experiments show that \textit{DE-RRD} outperforms the state-of-the-art competitors and achieves comparable or even better performance to that of the teacher model with faster inference~time.
\end{abstract}



\author{SeongKu Kang, Junyoung Hwang, Wonbin Kweon, Hwanjo Yu\*}
\affiliation{%
   \institution{Dept. of Computer Science and Engineering, POSTECH, South Korea}
   \{seongku, jyhwang, kwb4453, hwanjoyu\}@postech.ac.kr
}
\authornotemark[0]
\authornote{Corresponding Author}

\maketitle

\section{Introduction}
\label{sec:introduction}

In recent years, recommender system (RS) has been broadly adopted in various industries, helping users’ decisions in the era of information explosion, and playing a key role in promoting corporate profits.
However, a growing scale of users (and items) and sophisticated model architecture to capture complex patterns make the size of the model continuously increasing \cite{RD, CD, GCN_distill, DCF}.
A large model with numerous parameters has a high capacity, and thus usually has better recommendation performance.
On the other hand, it requires a large computational time and memory costs, and thus incurs a high latency during the inference phase, which makes it difficult to apply such large model to real-time platform.

Motivated by the significant success of \textit{knowledge distillation} (KD) in the computer vision field, a few work \cite{RD, CD} have employed KD for RS to reduce the size of models while maintaining the performance.
KD is a model-agnostic strategy to accelerate the learning of a new compact model (student) by transferring knowledge from a previously trained large model (teacher) \cite{hinton2015distilling}.
The knowledge transfer is conducted as follows:
First, the teacher model is trained with the user-item interactions in the training set which has binary labels -- `1' for observed interactions, and `0' for unobserved interactions.
Then, the student model is trained with the ``soft'' labels generated by the teacher model (i.e., teacher's predictions) along with the available binary labels.
The student model trained with KD has comparable performance to that of the teacher, and also has a lower latency due to its small size \cite{RD, CD}.

The core idea behind this process is that the soft labels predicted by the teacher model reveal hidden relations among entities (i.e., users and items) not explicitly included in the training set, so that they accelerate and improve the learning of the student model.
Specifically, the items ranked near the top of a user’s recommendation list would have strong correlations to the items that the user has interacted before \cite{RD}.
Also, the soft labels provide guidance for distinguishing the items that each user would like and the items that each user would not be interested in among numerous unobserved items only labeled as `0' \cite{CD}.
By using the additional supervisions from the teacher model, the state-of-the-art methods \cite{RD, CD} have achieved comparable or even better performance to the teacher models with faster inference time.

\begin{figure}[t]
  \includegraphics[width=0.49\textwidth]{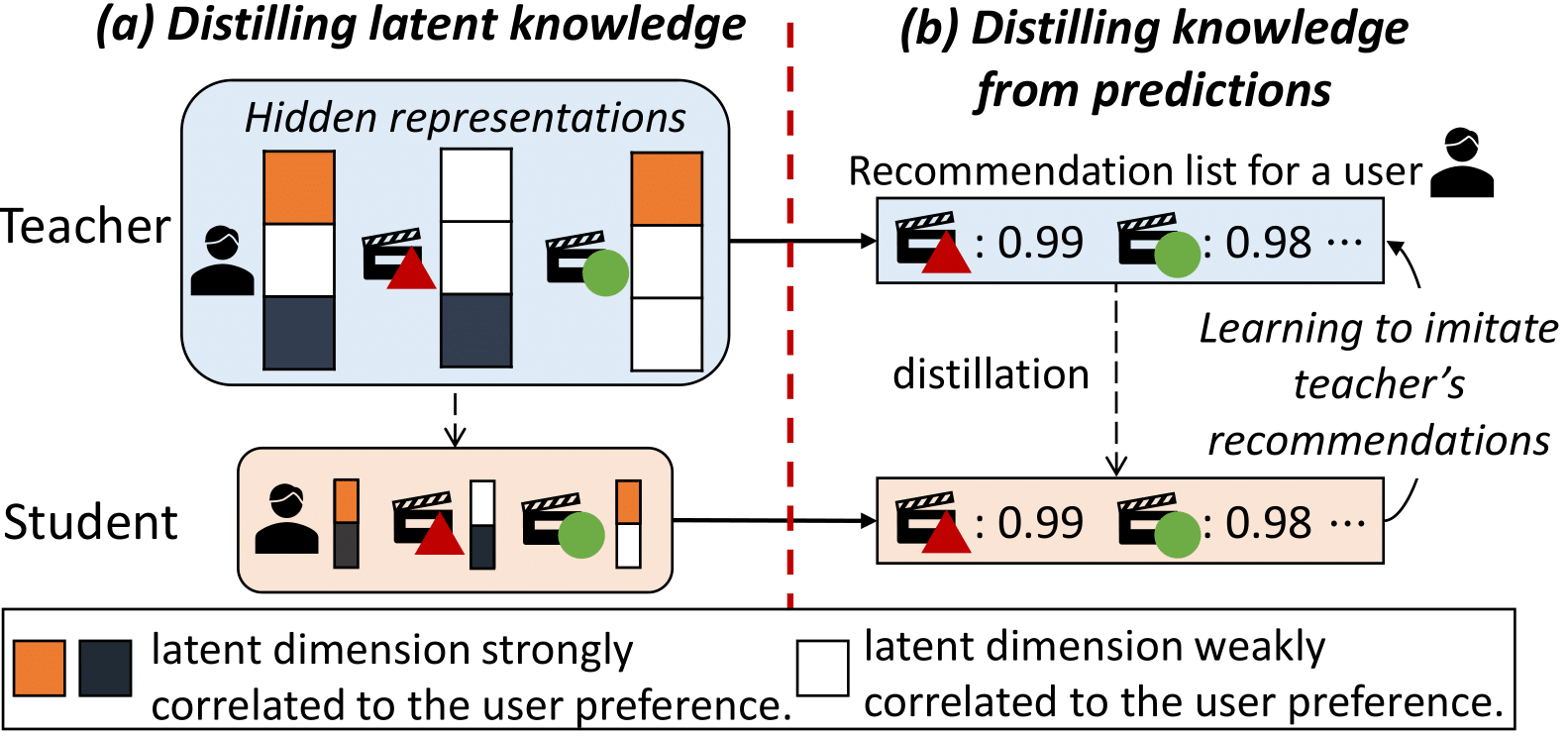}
  \caption{The existing methods \cite{RD, CD} distill the knowledge only based on the teacher’s predictions (b). The proposed framework directly distills the latent knowledge stored in the teacher (a) along with the knowledge revealed from the predictions (b).}
  \label{fig:overview}
\end{figure}

However, there are still limitations in existing methods \cite{RD, CD}.
First, the learning of the student is only guided by the teacher's prediction results, which is not sufficient to fully take advantage of the knowledge stored in the teacher.
This is because the prediction results incompletely reveal the teacher’s knowledge.
As illustrated in Figure 1, the recommendation list from the teacher only shows that a user has a similar degree of preference on the two items (0.99 and 0.98).
However, \textit{latent knowledge} in the teacher, which is used to make such predictions, contains more detailed information that the user likes different aspects of the two items (marked as navy blue and orange, respectively).
In this regard, we argue that the training process and the performance of the student can be further improved by directly distilling such latent knowledge stored in the teacher model.
Second, they distill the knowledge from the teacher's predictions in a point-wise manner that considers a single item at a time.
Because the point-wise approach does not consider multiple items simultaneously, it has a limitation in accurately maintaining the ranking orders predicted by the teacher model \cite{NCR}, which leads to degraded recommendation performance.

In this paper, we propose a novel knowledge distillation framework for RS, named DE-RRD, which distills both the latent knowledge stored in the teacher model (Fig. \ref{fig:overview}a) and the knowledge revealed from teacher's predictions (Fig. \ref{fig:overview}b).
By learning both the teacher's final predictions and the detailed knowledge that provides the bases for such predictions, the student model can be further improved.
The proposed framework consists of two methods: 1) Distillation Experts (DE) and 2) Relaxed Ranking Distillation (RRD).
The main contributions of this paper lie in the following aspects:

\vspace{3pt} \noindent
\textbf{Distilling latent knowledge in the teacher model.}
We propose a novel method---\textbf{DE}---for directly distilling latent knowledge stored in the teacher model.
Specifically, DE transfers the knowledge from hidden representation space (i.e., the output of the intermediate layer) of the teacher to the representation space of the student.
Due to the limited capacity, the student model cannot learn all the knowledge in the teacher representation space.
DE first introduces an ``expert'', which is a small feed-forward network, to distill the summarized knowledge that can restore the detailed knowledge of each entity in the teacher.
However, distilling the knowledge of all entities with a single expert intermingles the information of weakly correlated entities and further hinders the entire distillation process.
To tackle this problem, DE adopts the multiple experts and a novel expert selection strategy that clearly distinguishes the knowledge that each expert distills based on the correlations among the entities in the teacher representation space.
To the best of our knowledge, our approach is the first attempt to directly distill the latent knowledge in the teacher model for RS.
We demonstrate its rationality and superiority through extensive experiments and comprehensive analyses.

\vspace{1.5pt} \noindent
\textbf{Relaxed Ranking Distillation from the teacher’s predictions.}
We propose a new method---\textbf{RRD}---that transfers the knowledge from the teacher's predictions with direct consideration of ranking orders among items.
Unlike the existing methods \cite{RD, CD} that distill the knowledge of an item at a time, RRD formulates this as a ranking matching problem between the recommendation list of the teacher and that of the student.
To this end, RRD adopts the list-wise learning-to-rank approach \cite{xia2008list-wise} and learns to ensure the student to preserve the ranking orders predicted by the teacher.
However, directly applying the list-wise approach can have adverse effects on the recommendation performance.
Since a user is interested in only a few items among the numerous total items \cite{candidategeneration}, learning the detailed ranking orders of all items is not only daunting but also ineffective.
To tackle this challenge, RRD reformulates the daunting task to a \textit{relaxed ranking matching} problem.
Concretely, RRD matches the recommendation list from the teacher and that from the student, \textit{ignoring} the detailed ranking orders among the uninteresting items that the user would not be interested in.
RRD achieves superior recommendation performance compared to the state-of-the-art methods \cite{RD, CD}.



\vspace{2pt} \noindent
\textbf{An unified framework.}
We propose a novel framework---\textbf{DE-RRD}---which enables the student model to learn both from the teacher’s predictions and from the latent knowledge stored in the teacher model.
Our extensive experiments on real-world datasets show that DE-RRD considerably outperforms the state-of-the-art competitors. 
DE-RRD achieves comparable performance to that of the teacher with a smaller number of learning parameters than all the competitors.
Also, DE-RRD shows the largest performance gain when the student has the identical structure to the teacher model (i.e., self-distillation \cite{self_distill1}).
Furthermore, we provide both qualitative and quantitative analyses to further investigate the superiority of each proposed component.
The source code of DE-RRD is publicly available\footnote{\url{https://github.com/SeongKu-Kang/DE-RRD_CIKM20}}.

\section{Related Work}
\label{sec:relatedwork}
\label{reference}
Balancing \textit{effectiveness} and \textit{efficiency} is a key requirement for real-time recommender system (RS);
the system should provide \textit{accurate recommendations} with \textit{fast inference time}.
Recently, the size of the recommender model is continuously increasing, and the computational time and memory cost required for the inference are also increasing accordingly \cite{RD, CD, GCN_distill, DCF}.
Due to the high latency, it becomes difficult to apply such large recommender to the real-time large-scale platform.
In this section, we review several approaches to alleviate this problem.

\vspace{2pt} \noindent
\textbf{Balancing Effectiveness and Efficiency.}
Several methods have adopted hash techniques to reduce the inference cost \cite{hash1, hash2, DCF, candidategeneration}.
They first learn binary representations of users and items, then construct the hash table.
Although exploiting the binary representation can significantly reduce the inference costs, due to the constrained capability, their recommendation performance is limited compared to models that use real-values representations.
In addition, several work has focused on accelerating the inference of the existing recommenders \cite{MIPS1, tree_RS, compression1}.
Specifically, tree-based data structures \cite{ KDtree}, data compression techniques \cite{compression1}, and approximated nearest neighbor search techniques \cite{LSH, LSH_inner_product} have been successfully adopted to reduce the  search costs.
However, they still have problems such as applicable only to specific models (e.g., k-d tree for metric learning-based models \cite{METAS}), or easily falling into a local optimum due to the local search.

\vspace{2pt} \noindent
\textbf{Knowledge Distillation.}
Knowledge distillation (KD) is a model-agnostic strategy to improve the learning and the performance of a new “compact” model (student) by transferring knowledge from a previously trained “large” model (teacher) \cite{hinton2015distilling, romero2014fitnets, chen2017learning, self_distill1}.
The student model trained with KD has comparable performance to that of the teacher model, and also has lower inference latency due to its ~small size.
Most KD methods have focused on the image classification problem.
An early work \cite{hinton2015distilling} matches the softmax distribution of the teacher and the student.
The predicted label distribution contains more rich information (e.g., inter-class correlation) than the one-hot class label, which leads to improved learning of the student model.
Subsequent methods \cite{romero2014fitnets, chen2017learning} have focused on distilling knowledge from intermediate layers.
Because teacher's intermediate layers are generally bigger than that of the student, they \cite{romero2014fitnets, chen2017learning} utilize additional layers to bridge the different dimensions.
Interestingly, KD has turned out to be effective in improving the teacher model itself by self-distillation \cite{self_distill1}.

\vspace{2pt} \noindent
\textbf{Knowledge Distillation in Recommender System.}
Recently, inspired by the huge success of KD in the computer vision field, a few work \cite{RD, CD} have adopted KD to RS.
A pioneer work is Ranking Distillation (RD) \cite{RD} which applies KD for the ranking problem; Providing recommendations of top-$N$ unobserved items that have not interacted with a user.
RD jointly optimizes a base recommender's loss function with a \textit{distillation loss}.
\begin{equation}
    \begin{aligned}
\min_{\theta_{s}} \mathcal{L}_{B a s e} + \lambda \mathcal{L}_{R D}
    \end{aligned}
\end{equation}
where $\theta_{s}$ is the learning parameters of the student model, $\lambda$ is a hyperparameter that controls the effects of RD.
The base recommender can be any existing RS model such as BPR \cite{BPR}, NeuMF \cite{NeuMF}, and $\mathcal{L}_{B a s e}$ is its loss function (e.g., binary cross-entropy).
The distillation loss of RD for user $u$ is defined as follows:
\begin{equation}
    \begin{aligned}
    \mathcal{L}_{R D} = - \sum_{\pi_k \in \boldsymbol{\pi}} w_{\pi_k} \log \bigl( P\left(rel=1 | u, \pi_k \right)  \bigr)
    \end{aligned}
\end{equation}
where $\boldsymbol{\pi}$ is a ranked list of top-$K$ unobserved items for user $u$ predicted by the teacher, $\pi_k$ is the $k$-th item in this ranking, and $P\left(rel=1 | u, \pi_k \right)$ is the relevance probability of user $u$ to $\pi_k$ predicted by the student model. 
$w_{\pi_k}$ is the weight, which is computed based on each item's ranking from the student and the teacher, for reflecting relative importance among top-$K$ items.

A subsequent work
Collaborative Distillation (CD) \cite{CD} first samples unobserved items from the teacher's recommendation list according to their ranking; high-ranked items are more frequently sampled, then trains the student to mimic the teacher's prediction score (e.g., relevance probability) on the sampled items.
The distillation loss of CD for user $u$ is defined as follows:
\begin{equation}
    \begin{aligned}
\mathcal{L}_{C D} = - & \Bigl( \sum_{\pi_k \in \boldsymbol{\pi}} q_{\pi_k} \log \bigl( P\left(rel=1 | u, \pi_k \right)  \bigr)  \\ &+ (1-q_{\pi_k}) \log \bigl( 1-P\left(rel=1 | u, \pi_k \right)  \bigr) \Bigr)
    \end{aligned}
\end{equation}
where $\boldsymbol{\pi}$ is a ranked list of $K$ unobserved items sampled from teacher's recommendations for user $u$, $q_{\pi_k}$ is the weight, which is computed based on teacher's prediction score on each item, for reflecting relative importance among the sampled items.

In summary, the distillation loss of the existing methods makes the student model follow the teacher's predictions on unobserved items with particular emphasis on the high-ranked items.
In RS, only high-ranked items in the recommendation list are matter.
Also, such high-ranked items reveal hidden patterns among entities (i.e., users and items); the high-ranked items in the recommendation list would have strong correlations to the user \cite{RD}.
By using such additional supervisions from the teacher, they have achieved the comparable performance to the teacher with faster inference time.

However, the existing methods still have room for improvement by the following reasons:
First, the student can be further improved by directly distilling the \textit{latent knowledge} stored in the teacher model.
Latent knowledge refers to all information of users, items, and relationships among them that is discovered and stored in the teacher model.
Such knowledge is valuable for the student because it provides detailed explanations on the final prediction of the teacher.
Second, they transfer the knowledge from the teacher's predictions with a point-wise approach that considers a single item at a time.
Since the point-wise approach does not take into account multiple items simultaneously, it has a limitation in accurately maintaining the ranking orders in the teacher's ranking list \cite{NCR}.
This can lead to limited recommendation performance.

\section{Problem Formulation}
\label{sec:problem}
In this work, we focus on top-$N$ recommendations for implicit feedback.
Let $\mathcal{U}$ and $\mathcal{I}$ denote the set of users and items, respectively.
Given collaborative filtering (CF) information (i.e., implicit interactions between users and items), we build a binary matrix $\boldsymbol { R } \in \{0,1\}^{| \mathcal { U } | \times | \mathcal { I } |}$. 
Each element of $\boldsymbol { R }$ has a binary value indicating whether a user has interacted with an item (1) or not (0).
Note that an unobserved interaction does not necessarily mean a user's negative preference on an item, it can be that the user is not aware of the item.
For each user, a recommender model ranks all items that have not interacted with the user (i.e., unobserved items) and provides a ranked list of top-$N$ unobserved items.

The knowledge distillation is conducted as follows:
First, a teacher model with a large number of learning parameters is trained with the training set which has binary labels.
Then, a student model with a smaller number of learning parameters is trained with the help from the teacher model in addition to the binary labels.
The goal of KD is to fully improve the inference efficiency without compromising the effectiveness;
We aim to design a KD framework that enables the student model to maintain the recommendation performance of the teacher with a small number of learning parameters.

\section{DE-RRD: The Proposed framework}
\label{sec:method}
\begin{figure*}[t]
  \includegraphics[width=0.95\textwidth]{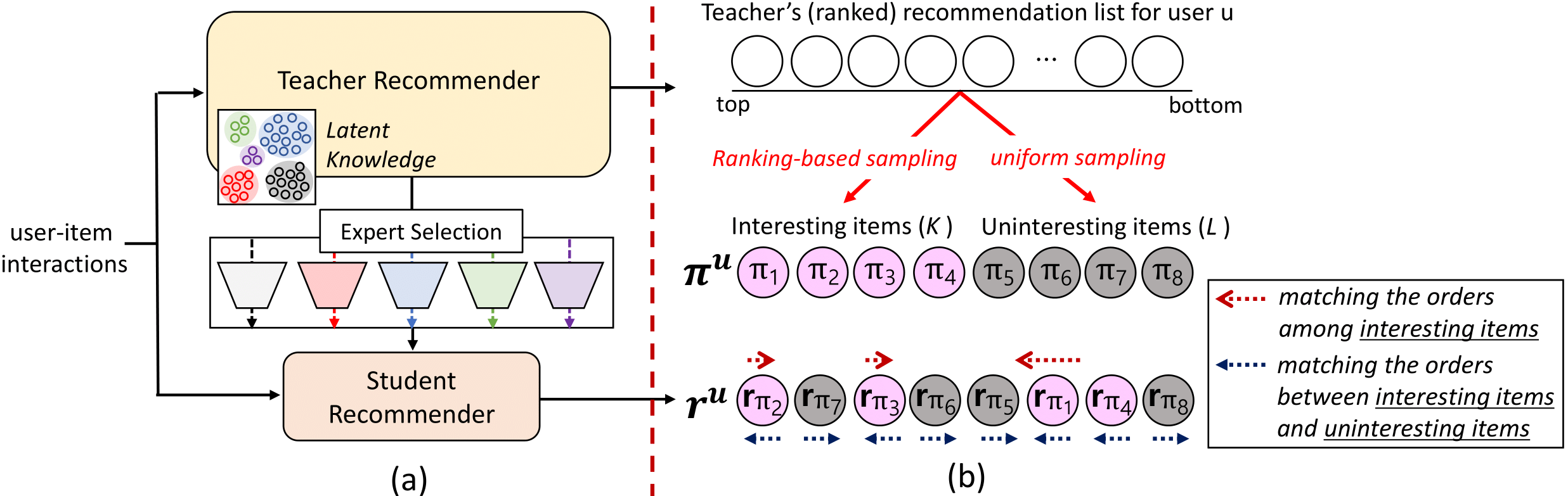}
  \caption{Illustration of DE-RRD framework.  (a) \textit{Distillation Experts} (DE) directly distills the teacher's latent knowledge with the experts and the selection strategy. (b) \textit{Relaxed Ranking Distillation} (RRD) distills the knowledge from the teacher's prediction based on the relaxed ranking approach that ignores  orders among the uninteresting items. Best viewed in color.}
  \label{fig:method}
  \vspace{-0.3cm}
\end{figure*}

We propose DE-RRD framework which enables the student model to learn both from the teacher’s predictions and from the latent knowledge encoded in the teacher model.
DE-RRD consists of two methods: 1) \textit{Distillation Experts} (DE) that directly transfers the latent knowledge from the teacher, 2) \textit{Relaxed Ranking Distillation} (RRD) that transfers the knowledge revealed from the teacher’s predictions with direct consideration of ranking orders among items. 
This section is organized as follows.
We first describe each component of the proposed framework: DE in Section 4.1, RRD in Section 4.2.
Then, we explain the end-to-end optimization process in Section 4.3. 
The overview of DE-RRD is provided in Figure \ref{fig:method}.

\vspace{-0.2cm}
\subsection{Distillation Experts (DE)}
In this section, we provide the details of DE which distills the latent knowledge from the hidden representation space (i.e., the output of the intermediate layer) of the teacher to the corresponding representation space of the student.
We first introduce ``expert'' to distill the summarized knowledge that can restore the detailed teacher's knowledge of each entity.
Then, we introduce a novel expert selection strategy for effectively distilling CF knowledge that contains information of all the entities having diverse preferences and characteristics.

\subsubsection{\textbf{Expert for distillation.}}
DE exploits “expert” to distill knowledge from the teacher's hidden representation space.
An expert, which is a small feed-forward network, is trained to \textit{reconstruct} the representation on a selected intermediate layer of the teacher from the representation on the corresponding intermediate layer of the student.
Let $h_t(\cdot)$ denote a mapping function to the representation space ($\in \mathbb{R}^{d_t}$) of the teacher model (i.e., a nested function up to the intermediate layer of the teacher).
Similarly, let $h_s(\cdot)$ denote a mapping function to the student’s representation space ($\in \mathbb{R}^{d_s}$).
The output of the mapping function can be a separate representation of a user, an item (e.g., BPR \cite{BPR}) or their combined representation (e.g., NeuMF \cite{NeuMF}) based on the base model’s structure and the type of selected layer.
Here, we use user $u$ as an example for convenience.
An expert $E$ is trained to reconstruct $h_{t}\left(u\right)$ from $h_{s}\left(u\right)$ as follows:
\begin{equation}
\begin{aligned}
\mathcal{L}(u)=\| h_{t}\left(u\right) - E \bigl(h_{s}\left(u\right)\bigr)\|_2
\end{aligned}
\end{equation}
Note that in the KD process, the teacher model is already trained and frozen.
By minimizing the above equation, parameters in the student model (i.e., $h_s(\cdot)$) and the expert are updated.

The student model has smaller capacity compared to the teacher ($d_s << d_t$). 
By minimizing the equation 4, the student learns compressed information on the user's preference that can restore more detailed knowledge in the teacher as accurate as possible.
This approach provides a kind of filtering effect and improves the learning of the student model.

\subsubsection{\textbf{Expert selection strategy.}}
Training a single expert to distill all the CF knowledge in the teacher is not sufficient to achieve satisfactory performance.
The CF knowledge contains vast information of user groups with various preferences and item groups with diverse characteristics.
When a single expert is trained to distill the knowledge of all the diverse entities, the information of the weakly correlated entities (e.g., users that have dissimilar preferences) is mixed and reflected in the expert's weights.
This leads to the adulterated distillation that hinders the student model from discovering some users' preferences.

To alleviate the problem, DE puts multiple experts in parallel and clearly distinguishes the knowledge that each expert distills.
The key idea is to divide the representation space into exclusive divisions based on the teacher's knowledge and make each expert to be specialized in distilling the knowledge in a division (Fig. \ref{fig:method}a).
The representations belonging to the same division has strong correlations with each other, and they are distilled by the same expert without being mixed with weakly correlated representations belonging to the different divisions.
The knowledge transfer of DE is conducted in the two steps:
1) a selection network first computes each expert's degree of specialization for the knowledge to be distilled.
2) DE selects an expert based on the computed distribution, then distills the knowledge through the selected expert.

Concretely, DE has $M$ experts ($E_1, E_2, ... , E_M$) and a selection network $S$ whose output is $M$-dimensional vector.
To distill user $u$'s knowledge from the teacher, the selection network $S$ first computes the normalized specialization score vector $\boldsymbol{\alpha}^{u} \in \mathbb{R}^{M}$ as follows:
\begin{equation}
\begin{aligned}
\mathbf{e}^{u} &= S\bigl(h_{t}\left(u\right)\bigr),\\
\alpha^{u}_{m} &= \frac{\exp \left(e^{u}_m\right)}{\sum_{i=1}^{M} \exp(e^{u}_i)} \quad \text{for} \quad m = 1, ..., M
\end{aligned}
\end{equation}

\noindent
Then, DE selects an expert based on the computed distribution.
We represent the selection variable $\mathbf{s}^{u}$ that determines which expert to be selected for distilling $h_t(u)$.
$\mathbf{s}^{u}$ is a $M$-dimensional one-hot vector where an element is set to 1 if the corresponding expert is selected for distillation.
DE samples this selection variable $\mathbf{s}^u$ from a multinoulli distribution parameterized by $\{\alpha^{u}_m\}$ i.e., $p\left(s^{u}_{m}=1 | S, h_{t}\left(u\right)\right) = \alpha^{u}_{m}$, then reconstructs teacher's representation as follows:
\begin{equation}
\begin{aligned}
\mathbf{s}^{u} &\sim \text{Multinoulli}_M \left(\{\alpha^{u}_{m}\}\right)\\
\mathcal{L}(u) &=\|h_{t}\left(u\right)-\sum_{m=1}^{M} s^{u}_m \cdot E_{m}\bigl(h_{s}\left(u\right)\bigr)\|_{2}
\end{aligned}
\end{equation}
\noindent
However, the sampling process is non-differentiable, which would block the gradient flows and disable the end-to-end training. 
As a workaround, we adopt a continuous relaxation of the discrete distribution by using Gumbel-Softmax \cite{GumbelSoftmax}. 
The Gumbel-Softmax is a continuous distribution on the simplex that can approximate samples from a categorical distribution; it uses the Gumbel-Max trick \cite{GumbelMax} to draw samples from the categorical distribution, then uses the softmax function as a continuous approximation of argmax operation to get the approximated one-hot representation.
With the relaxation, the selection network can be trained by the backpropagation.

\begin{figure}[t]
  \includegraphics[width=0.45\textwidth]{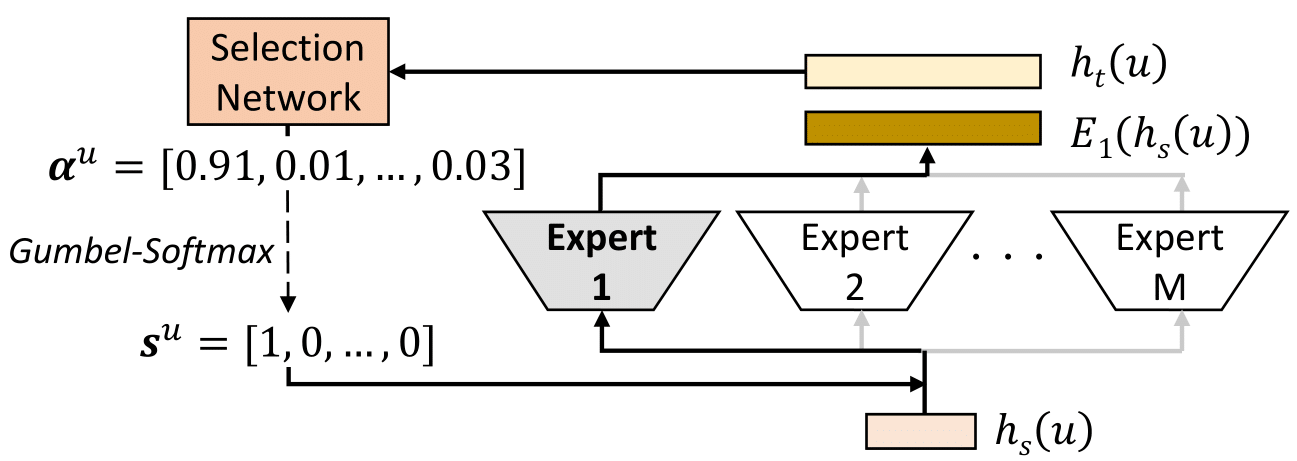}
  \caption{Illustration of the expert selection process of DE. During the training, $\mathbf{s}^{u}$ becomes a one-hot vector and selects the most specialized expert in the knowledge to be distilled.}
  \label{fig:selection}
\end{figure}

DE gets the approximated one-hot selection variable $\mathbf{s}^{u}$ by using the Gumbel-Softmax and reconstructs the teacher's representation as~follows:
\begin{equation}
\begin{aligned}
s^{u}_{m} &= \frac{\exp \Bigl(\left(\log \alpha^{u}_m +g_{m} \right) / \tau \Bigr)}{\sum_{i=1}^{M} \exp \Bigl(\bigl(\log \alpha^{u}_i +g_{i}\bigr)/ \tau\Bigr)}  \quad \text{for} \quad m = 1, ..., M\\
&\mathcal{L}(u) =\|h_{t}\left(u\right)-\sum_{m=1}^{M} s^{u}_m \cdot E_{m}\bigl(h_{s}\left(u\right)\bigr)\|_{2}
\end{aligned}
\end{equation}
where $g_i$ is i.i.d drawn from Gumbel$(0, 1)$ distribution\footnote{$g_i=-\text{log}(-\text{log}(r))$, where $r$ is sampled from $Uniform(0,1)$.}.
The extent of relaxation is controlled by a temperature parameter $\tau$.
In the beginning of the training, we set a large value on $\tau$, and gradually decreases its value during the training.
As $\tau$ is decreased to $0$, 
$\mathbf{s}^{u}$ smoothly becomes one-hot vector where $s^{u}_m = 1$ with probability $\alpha^{u}_m$.
In other words, during the training, each expert gradually gets specialized on certain information that has strong correlations.
This process is illustrated in Figure \ref{fig:selection}.

\vspace{1pt} \noindent
\textbf{Discussion: Effects of expert selection.}
As the expert selection is based on the teacher's knowledge, correlations among the entities in the teacher representation space are naturally reflected in the expert selection;
the user representations with very similar preferences (i.e., located closely in the space) would be distilled by the same expert with a high probability.
This allows each expert to be trained to distill only the knowledge of strongly correlated entities, and thus each expert can provide better guidance that does not include the information of weakly correlated entities.

\vspace{1pt} \noindent
\textbf{Discussion: selection vs. attention.}
Instead of selecting one expert, the attention mechanism (i.e., the softmax function) can be adopted.
However, we think the selection is a more appropriate choice to distill the CF knowledge containing all the entities having diverse preferences and characteristics.
This is because the attention makes every expert involved in distilling the knowledge of each entity.
In other words, like in the case of a single expert, all the experts and attention network are trained to minimize the overall reconstruction errors of all the diverse entities. 
By doing so, information of weakly relevant entities gets mixed together, and this leads to performance degrade in some user groups.
We provide experiment results to support our claims. Please refer to Section~5.3.

\vspace{-0.2cm}
\subsubsection{\textbf{Optimization of DE}}
DE is jointly optimized with the base model's loss function in the end-to-end manner as follows:
\begin{equation}
\begin{aligned}
\min_{\theta_{s}, \theta_{DE}} \mathcal{L}_{B a s e} + \lambda_{D E} \cdot \mathcal{L}_{D E}
\end{aligned}
\end{equation}
where $\theta_{s}$ is the learning parameters of the student model, $\theta_{DE}$ is the learning parameters of DE (i.e., the selection network and the experts), and $\lambda_{D E}$ is a hyperparameter that controls the effects of DE.
The base model can be any existing recommender (e.g., BPR, NeuMF), and $\mathcal{L}_{B a s e}$ corresponds to its loss function.
Note that the experts are not used in the inference phase.

The loss function of DE can be flexibly defined based on the base model's structure and the types of hidden layer chosen for the distillation.
Concretely, for NeuMF \cite{NeuMF}, which is a state-of-the-art deep recommender, the loss function can be defined to 1) separately distill knowledge of users and items in a mini-batch (i.e., $\sum_{u \in B} \mathcal{L}(u) +  \sum_{i \in B} \mathcal{L}(i)$) 
or 2) distill the combined knowledge (i.e., $\sum_{(u, i) \in B} \mathcal{L}(u, i)$). 
Also, we adopt a simple temperature annealing schedule, which gradually decays the temperature from $\tau_0$ to $\tau_P$ as done in \cite{DRE}:
$\tau (p)=\tau_0(\tau_P / \tau_0)^{p/P}$ where $\tau(p)$ is the temperature at epoch $p$, and $P$ is the total training epochs.

\subsection{Relaxed Ranking Distillation (RRD)}
We propose RRD, a new method to distill the knowledge revealed from the teacher's predictions with direct consideration of ranking orders among items.
RRD formulates this as a ranking matching problem between the recommendation list of the teacher model and that of the student model.
To this end, RRD adopts the classical list-wise learning-to-rank approach \cite{xia2008list-wise}.
Its core idea is to define a probability of a permutation (i.e., a ranking order) based on the ranking score predicted by a model, and train the model to maximize the likelihood of the ground-truth ranking order.
For more details about the list-wise approach, please refer to \cite{xia2008list-wise}.

However, merely adopting the list-wise loss can have adverse effects on the ranking performance.
Because a user is interested in only a few items among the numerous total items \cite{candidategeneration}, learning the detailed ranking orders of all the unobserved items is not only daunting but also ineffective.
The recommendation list from the teacher model contains information about a user’s potential preference on each unobserved item; A few items that the user would be interested in (i.e., interesting items) are located near the top of the list, whereas the majority of items that the user would not be interested in (i.e., uninteresting items) are located far from the top.

Based on this information, RRD reformulates the daunting task of learning all the precise ranking orders to a \textit{relaxed ranking matching} problem.
In other words, RRD aims to match the recommendation list from the teacher and that from the student, \textit{ignoring} the detailed ranking orders among the uninteresting items.
Concretely, RRD distills the information of 1) the detailed ranking orders among the interesting items, 2) the relative ranking orders between the interesting items and the uninteresting items.
The overview of RRD is provided in Figure 2b.

\subsubsection{\textbf{Sampling interesting/uninteresting items.}}
The first step of RRD is to sample items from the teacher's recommendation list.
In specific, RRD samples $K$ interesting items and $L$ uninteresting items for each user.
As a user would not be interested in the vast majority of items, the interesting items should be sampled from a very narrow range near the top of the list, whereas the uninteresting items should be sampled from the wide range of the rest.
To sample the interesting items, we adopt a ranking position importance scheme \cite{rendle2014improving, RD} that places more emphasis on the higher positions in the ranking list.
In the scheme, the probability of the $k$-th ranked item to be sampled is defined as: $p_k \propto e^{-k/T}$
where $T$ is the hyperparameter that controls emphasis on top positions.
With the scheme, RRD samples $K$ interesting items according to the user’s potential preference on each item (i.e., item's ranking) predicted by the teacher. 
To sample the uninteresting items that corresponds the majority of items, we use a simple uniform sampling.
Concretely, RRD uniformly samples $L$ uninteresting items from a set of items that have lower rankings than the previously sampled interesting~items.

\subsubsection{\textbf{Relaxed permutation probability.}}
Then, RRD defines a relaxed permutation probability motivated by \cite{xia2008list-wise}.
For user $u$, $\boldsymbol{\pi^{u}}$ denotes a ranked list of all the sampled items ($K+L$) sorted by the original order in the teacher's recommendation list.
$\mathbf{r}^{u}$ denotes ranking scores on the sampled items predicted by the student model.
The relaxed permutation probability is formulated as follows:
\begin{equation}
\begin{aligned}
    p\left(\boldsymbol{\pi}^u_{1:K} | \mathbf{r}^u\right)=\prod_{k=1}^{K} \frac{\exp ({r}^u_{\pi_{k}})}{\sum_{i=k}^{K} \exp ({r}^u_{\pi_{i}})+\sum_{j=K}^{K+L} \exp ({r}^u_{\pi_{j}})}
\end{aligned}
\end{equation}
where ${r}^{u}_{\boldsymbol{\pi}_k}$ denotes a ranking score predicted by the student for the $k$-th item in $\boldsymbol{\pi}^u$, $\boldsymbol{\pi}^u_{1:K}$ denotes the partial list that contains the interesting items.
RRD learns to maximize the log-likelihood $\log p\left(\boldsymbol{\pi}_{1:K} | \mathbf{r}\right)$ for all users.
The proposed permutation probability is not affected by the detailed ranking orders among the uninteresting items ($L$).
By maximizing the log-likelihood, the student model is trained to locate all the interesting items ($K$) higher than all the uninteresting items ($L$) in the recommendation list, while maintaining the detailed ranking orders (from the teacher's recommendation list) among the interesting items.

\subsubsection{\textbf{Optimization of RRD}}
RRD is jointly optimized with the base model's loss function in the end-to-end manner as follows:
\begin{equation}
\begin{aligned}
\min_{\theta_{s}} \mathcal{L}_{B a s e} + \lambda_{R R D} \cdot \mathcal{L}_{R R D}
\end{aligned}
\end{equation}
where $\theta_{s}$ is the learning parameters of the student model and $\lambda_{R R D}$ is a hyperparameter that controls the effects of RRD.
The base model can be any existing recommender, and $\mathcal{L}_{B a s e}$ corresponds to its loss function.
The sampling process is conducted at every epoch.
The loss function of RRD is defined to distill the knowledge of users in the mini-batch: $-\frac{1}{|B|}\sum_{u \in B} \log p(\boldsymbol{\pi}^{u}_{1:K}|\mathbf{r}^{u})$.

\subsection{Optimization of DE-RRD}
The proposed DE-RRD framework is optimized in the end-to-end manner as follows:
\begin{equation}
\begin{aligned}
\min_{\theta_{s}, \theta_{DE}} \mathcal{L}_{B a s e} + \lambda_{D E} \cdot \mathcal{L}_{D E} + \lambda_{R R D} \cdot \mathcal{L}_{R R D}
\end{aligned}
\end{equation}
where $\theta_{s}$ is the learning parameters of the student model, $\theta_{DE}$ is the learning parameters of DE (i.e., the selection network and the experts).
The base model can be any existing recommender, and $\mathcal{L}_{B a s e}$ corresponds to its loss function.

\section{Experiments}
\label{sec:experiments}
We validate the superiority of DE-RRD on 12 experiment settings (2 real-world datasets $\times$ 2 base models $\times$ 3 different student model sizes).
We first provide extensive experiment results supporting that DE-RRD outperforms the state-of-the-art competitors (Section 5.2).
We also provide both quantitative and qualitative analyses to verify the rationality and superiority of each proposed component (Section 5.3).
Lastly, we provide hyperparameter study (Section 5.4).

\subsection{Experimental Setup}
\noindent
\textbf{Datasets.}
We use two public real-world datasets: CiteULike \cite{wang2013collaborative}, Foursquare \cite{liu2017experimental}.
We remove users and items having fewer than five ratings for CiteULike, twenty ratings for Foursquare as done in \cite{BPR, NeuMF, SSCDR}.
Data statistics are summarized in Table \ref{tbl:statistic}.
\begin{table}[h]
\renewcommand{\arraystretch}{0.44}
  \caption{Data Statistics (after preprocessing)}
  \begin{tabular}{ccccc}
    \toprule
    Dataset & \#Users & \#Items & \#Interactions & Sparsity \\
    \midrule
    CiteULike & 5,220 & 25,182 & 115,142 & 99.91\% \\
    Foursquare & 19,466 & 28,594 & 609,655 & 99.89\% \\
    \bottomrule
  \end{tabular}
    \label{tbl:statistic}
\end{table}

\vspace{2pt} \noindent
\textbf{Base Models.}
We validate the proposed framework on base models that have different architectures and optimization strategies.
We choose a latent factor model and a deep learning model that are broadly used for top-$N$ recommendation with implicit feedback.
\begin{itemize}[leftmargin=*]
    \item \textbf{BPR \cite{BPR}}: 
    A learning-to-rank model for implicit feedback.
    It assumes that observed items are more preferred than unobserved items and optimizes Matrix Factorization (MF) with the pair-wise ranking loss function.
    \item \textbf{NeuMF \cite{NeuMF}}: The state-of-the-art deep model for implicit feedback. 
    NeuMF combines MF and Multi-Layer Perceptron (MLP) to learn the user-item interaction, and optimizes it with the point-wise objective function (i.e., binary cross-entropy).
\end{itemize}

\vspace{2pt} \noindent
\textbf{Teacher/Student.}
For each base model and dataset, we increase the number of learning parameters until the recommendation performance is no longer increased, and use the model with the best performance as Teacher model.
For each base model, we build three student models by limiting the number of learning parameters.
We adjust the number of parameters based on the size of the last hidden layer.
The limiting ratios ($\phi$) are \{0.1, 0.5, 1.0\}.
Following the notation of the previous work \cite{RD, CD}, we call the student model trained without the help of the teacher model (i.e., no distillation) as ``Student'' in this experiment sections.

\vspace{2pt} \noindent
\textbf{Comparison Methods.}
The proposed framework is compared with the following methods:
\begin{itemize}[leftmargin=*]
    \item \textbf{Ranking Distillation (RD) \cite{RD}}: A KD method for recommender system that uses items with the highest ranking from the teacher's predictions for distilling the knowledge.
    \item \textbf{Collaborative Distillation (CD) \cite{CD}}: The state-of-the-art KD method for recommender system.
    CD samples items from teacher's predictions based on their ranking, then uses them for distillation.
    As suggested in the paper, we use unobserved items only for distilling the knowledge.
\end{itemize}
Finally, \textbf{DE-RRD} framework consists of the following two methods:
\begin{itemize}[leftmargin=*]
    \item \textbf{Distillation Experts (DE)}: A KD method that directly distills the latent knowledge stored in the teacher model.
    It can be combined with any \textit{prediction-based} KD methods (e.g., RD, CD, RRD).
    \item \textbf{Relaxed Ranking Distillation (RRD)}: A KD method that distills the knowledge revealed from the teacher’s predictions with consideration of relaxed ranking orders among items. 
\end{itemize}

\noindent
\textbf{Evaluation Protocol.}
We follow the widely used \textit{leave-one-out} evaluation protocol ~\cite{NeuMF, transCF, SSCDR}.
For each user, we leave out a single interacted item for testing, and use the rest for training.
In our experiments, we leave out an additional interacted item for the validation.
To address the time-consuming issue of ranking all the items, we randomly sample 499 
items from a set of unobserved items of the user, then evaluate how well each method can rank the test item higher than these sampled unobserved items. 
We repeat this process of sampling a test/validation item and unobserved items five times and report the average results.

As we focus on the top-$N$ recommendation task based on implicit feedback, we evaluate the performance of each method with widely used three ranking metrics \cite{NeuMF, SSCDR, candidategeneration}: 
hit ratio (H@$N$), normalized discounted cumulative gain (N@$N$), and mean reciprocal rank (M@$N$). 
H@$N$ measures whether the test item is present in the top-$N$ list, while N@$N$ and M@$N$ are position-aware ranking metrics that assign higher scores to the hits at upper ranks.

\vspace{2pt} \noindent
\textbf{Implementation Details for Reproducibility.}
We use PyTorch to implement the proposed framework and all the baselines, and use Adam optimizer to train all the methods.
For RD, we use the public implementation provided by the authors.
For each dataset, hyperparameters are tuned by using grid searches on the validation set. 
The learning rate for the Adam optimizer is chosen from \{0.1, 0.05, 0.01, 0.005, 0.001, 0.0005, 0.0001\}, the model regularizer is chosen from $\{10^{-1}, 10^{-2}, 10^{-3}, 10^{-4}, 10^{-5}\}$.
We set the total number of epochs as 1000, and adopt early stopping strategy; stopping if H@$5$ on the validation set does not increase for 30 successive epochs.
For all base models (i.e., BPR, NeuMF), the number of negative sample is set to 1, and no pre-trained technique is used.
For NeuMF, the number of the hidden layers is chosen from \{1, 2, 3, 4\}.

For all the distillation methods (i.e., RD, CD, DE, RRD), weight for KD loss ($\lambda$) is chosen from $\{1, 10^{-1}, 10^{-2}, 10^{-3}, 10^{-4}, 10^{-5}\}$.
For DE, the number of experts ($M$) is chosen from \{5, 10, 20, 30\}, MLP is employed for the experts and the selection network. 
The shape of the layers of an expert is [$d_s \rightarrow (d_s + d_t)/2 \rightarrow d_t$] with \textit{relu} activation, and that of the selection network is [$d_t \rightarrow M$].
We select the last hidden layer of all the base models to distill latent knowledge.
We put the experts according to the structure of the selected layer;
For the layer where user and item are separately encoded (i.e., BPR), we put $M$ user-side experts and $M$ item-side experts, and for the layer where user and items are jointly encoded (i.e., NeuMF), we put $M$ experts to distill the combined information.
$\tau_0$ and $\tau_{P}$ are set to $1, 10^{-10}$, respectively.
For prediction-based KD methods (i.e., RD, CD, RRD), the number of high-raked (or interesting) items ($K$) for distillation
is chosen from \{10, 20, 30, 40, 50\}, weight for controlling the importance of top position ($T$) is chosen from \{1, 5, 10, 20\}.
For RRD, the number of uninteresting items ($L$) is set to the same with $K$, but it can be further tuned.
For RD, the number of the warm-up epoch is chosen from \{30, 50, 100\}, the number of negative items in the dynamic weight is chosen from \{50, 100\}.
Also, RD and CD have additional hyperparameters for reflecting the relative importance of the items used for distillation.
We follow the recommended values from the public implementation and from the original papers.

\begin{table*}[t]
\setlength\tabcolsep{7pt}
\small
\RowStretch{0.2}
  \caption{Recommendation performances ($\phi=0.1$). \textit{Improv.b} and \textit{Improv.s} denote the improvement of DE-RRD over the best baseline and student respectively.
  *, **, ***, and **** indicate $p \leq 0.05$, $p \leq 0.005$, $p \leq 0.0005$, and $p \leq 0.00005$ for the paired t-test of vs. the best baseline (for RRD, DE-RRD), vs. Student (for DE) on H@5.}
  
  \begin{tabular}{cclccc ccc ccc}
    \toprule 
     Dataset & Base Model & KD Method & H@5 & M@5 & N@5 & H@10 & M@10 & N@10 & H@20 & M@20 & N@20 \\
    \midrule
     &&Teacher&0.5135&0.3583&0.3970&0.6185&0.3724&0.4310&0.7099&0.3788&0.4541\\
     &&Student&0.4441&0.2949&0.3319&0.5541&0.3102&0.3691&0.6557&0.3133&0.3906\\
     &&RD&0.4533&0.3019&0.3395&0.5601&0.3161&0.3740&0.6633&0.3232&0.3993\\
     &\multirow{6}{*}{BPR}&CD&0.4550&0.3025&0.3404&0.5607&0.3167&0.3746&0.6650&0.3240&0.4011\\
     \cmidrule{3-12}
     &&DE ** &0.4817&0.3230&0.3625&0.5916&0.3372&0.3977&0.6917&0.3441&0.4229\\
     &&RRD ** &0.4622&0.3076&0.3461&0.5703&0.3220&0.3809&0.6746&0.3293&0.4074\\
     &&DE-RRD ***&\textbf{0.4843}&\textbf{0.3231}&\textbf{0.3632}&\textbf{0.5966}&\textbf{0.3373}&\textbf{0.3989}&\textbf{0.6991}&\textbf{0.3447}&\textbf{0.4251}\\
     \cmidrule{3-12}
     &&\textit{Improv.b}&6.44\%&6.81\%&6.7\%&6.4\%&6.47\%&6.47\%&5.12\%&6.4\%&5.98\%\\
    \multirow{9}{*}{\rotatebox[origin=c]{0}{CiteULike}}&&\textit{Improv.s}&9.06\%&9.57\%&9.44\%&7.66\%&8.7\%&8.06\%&6.62\%&10.02\%&8.83\%\\
    \cmidrule{2-12}
    &&Teacher&0.4790&0.3318&0.3684&0.5827&0.3457&0.4020&0.6748&0.3521&0.4254\\
    &&Student&0.3867&0.2531&0.2865&0.4909&0.2670&0.3202&0.5833&0.2738&0.3436\\
    &&RD&0.4179&0.2760&0.3113&0.5211&0.2896&0.3444&0.6227&0.2958&0.3696\\
    &\multirow{6}{*}{NeuMF }&CD&0.4025&0.2633&0.2979&0.5030&0.2769&0.3306&0.6053&0.2822&0.3550\\
    \cmidrule{3-12}
    &&DE **&0.4079&0.2625&0.2986&0.5139&0.2766&0.3328&0.6238&0.2843&0.3607\\
    &&RRD ***&0.4737&0.3086&0.3497&0.5800&0.3236&0.3847&0.6765&0.3305&0.4094\\
    &&DE-RRD ****&\textbf{0.4758}&\textbf{0.3108}&\textbf{0.3518}&\textbf{0.5805}&\textbf{0.3246}&\textbf{0.3856}&\textbf{0.6770}&\textbf{0.3312}&\textbf{0.4099}\\
    \cmidrule{3-12}
    &&\textit{Improv.b}&13.83\%&12.6\%&13.03\%&11.42\%&12.09\%&11.95\%&8.72\%&11.95\%&10.9\%\\
    &&\textit{Improv.s}&23.03\%&22.79\%&22.8\%&18.26\%&21.58\%&20.42\%&16.07\%&20.95\%&19.28\%\\
    \midrule
     &&Teacher&0.5598&0.3607&0.4101&0.7046&0.3802&0.4571&0.8175&0.3882&0.4859\\
    &&Student&0.4869&0.3033&0.3489&0.6397&0.3239&0.3984&0.7746&0.3338&0.4333\\
    &&RD&0.4932&0.3102&0.3555&0.6453&0.3302&0.4045&0.7771&0.3391&0.4377\\
    &\multirow{6}{*}{BPR}&CD&0.5006&0.3147&0.3608&0.6519&0.3354&0.3237&0.7789&0.3440&0.4421\\
    \cmidrule{3-12}
    &&DE ****&0.5283&0.3344&0.3824&0.6810&0.3544&0.4316&0.8032&0.3631&0.4627\\
    &&RRD **&0.5132&0.3258&0.3722&0.6616&0.3455&0.4202&0.7862&0.3540&0.4516\\
    &&DE-RRD ****&\textbf{0.5308}&\textbf{0.3359}&\textbf{0.3843}&\textbf{0.6829}&\textbf{0.3565}&\textbf{0.4336}&\textbf{0.8063}&\textbf{0.3647}&\textbf{0.4647}\\
    \cmidrule{3-12}
     &&\textit{Improv.b}&6.03\%&6.74\%&6.51\%&4.76\%&6.29\%&7.19\%&3.52\%&6.02\%&5.11\%\\
    \multirow{9}{*}{\rotatebox[origin=c]{0}{Foursquare}}&&\textit{Improv.s}&9.02\%&10.75\%&10.15\%&6.75\%&10.06\%&8.84\%&4.09\%&9.26\%&7.25\%\\
    \cmidrule{2-12}
    &&Teacher&0.5436&0.3464&0.3954&0.6906&0.3662&0.4430&0.8085&0.3746&0.4731\\
    &&Student&0.4754&0.2847&0.3319&0.6343&0.3060&0.3833&0.7724&0.3157&0.4185\\
    &&RD&0.4789&0.2918&0.3380&0.6368&0.3110&0.3878&0.7761&0.3173&0.4205\\
    &\multirow{6}{*}{NeuMF}&CD&0.4904&0.2979&0.3456&0.6477&0.3156&0.3940&0.7845&0.3260&0.4293\\
    \cmidrule{3-12}
    &&DE *&0.4862&0.2977&0.3444&0.6413&0.3174&0.3938&0.7742&0.3278&0.4284\\
    &&RRD ***&0.5172&0.3110&0.3621&0.6739&0.3321&0.4132&0.7982&0.3409&0.4450\\
    &&DE-RRD ****&\textbf{0.5193}&\textbf{0.3130}&\textbf{0.3641}&\textbf{0.6741}&\textbf{0.3332}&\textbf{0.4139}&\textbf{0.7983}&\textbf{0.3421}&\textbf{0.4454}\\
    \cmidrule{3-12}
     &&\textit{Improv.b}&5.89\%&5.07\%&5.35\%&4.08\%&5.58\%&5.05\%&1.76\%&4.94\%&3.75\%\\
    &&\textit{Improv.s}&9.23\%&9.94\%&9.7\%&6.27\%&8.89\%&7.98\%&3.35\%&8.36\%&6.43\%\\
    \bottomrule
  \end{tabular}
  \label{tab:main}
  \vspace{-0.3cm}
\end{table*}

\begin{figure*}[t]
\begin{subfigure}[t]{0.5\linewidth}
    \includegraphics[width=\linewidth]{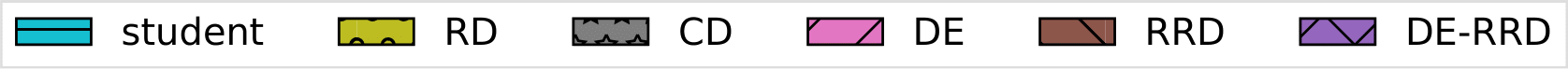}
\end{subfigure}

\begin{subfigure}[t]{0.243\linewidth}
    \includegraphics[width=\linewidth]{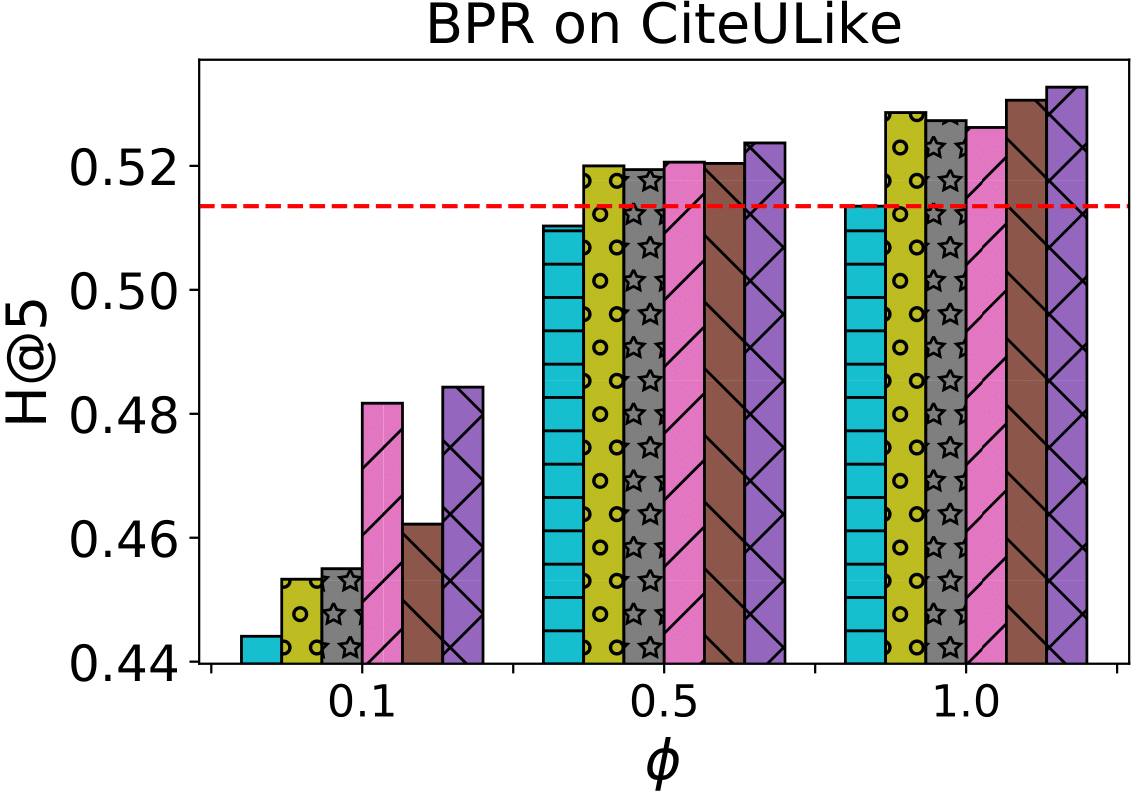}
\end{subfigure}
\begin{subfigure}[t]{0.243\linewidth}
    \includegraphics[width=\linewidth]{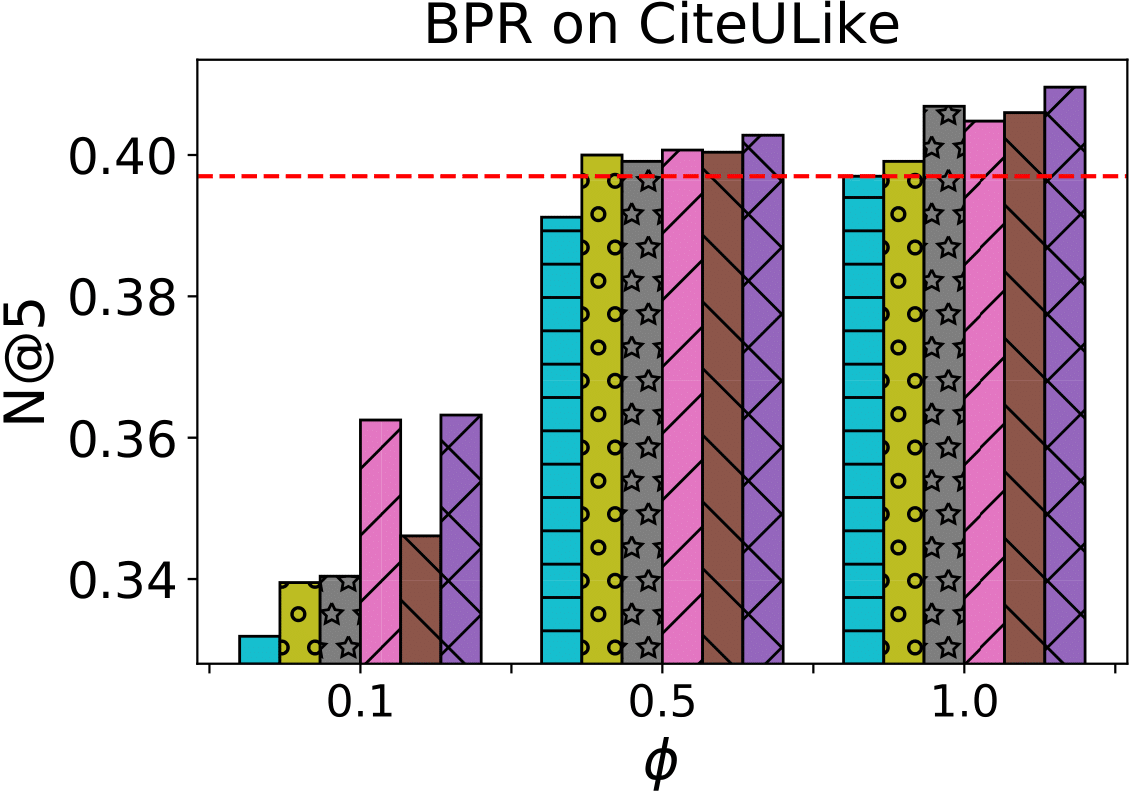}
\end{subfigure} 
\begin{subfigure}[t]{0.243\linewidth}
    \includegraphics[width=\linewidth]{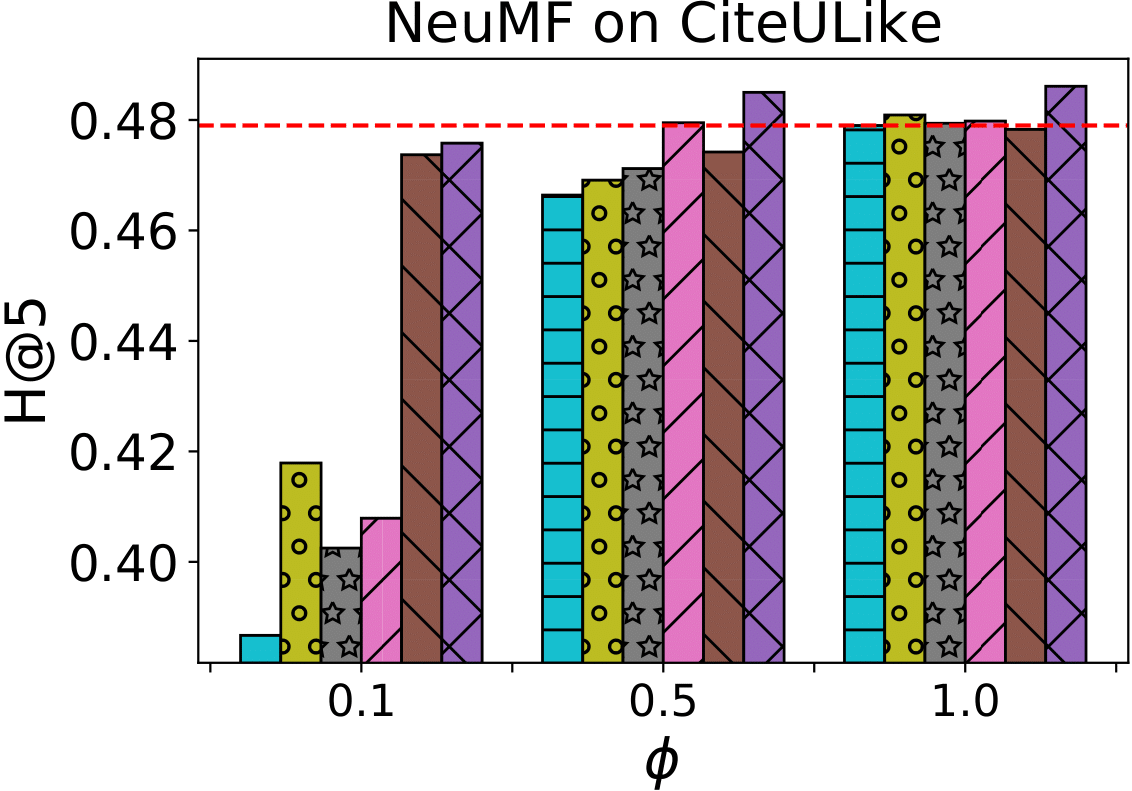}
\end{subfigure} 
\begin{subfigure}[t]{0.243\linewidth}
    \includegraphics[width=\linewidth]{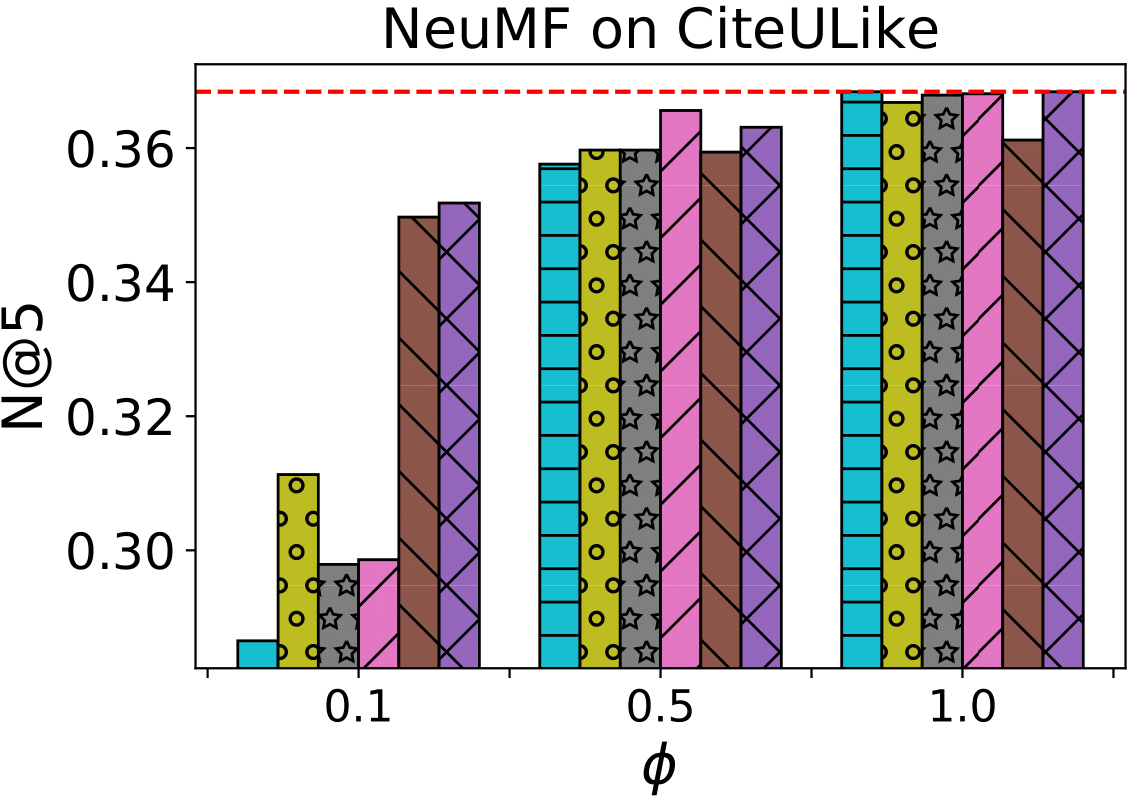}
\end{subfigure} 
\begin{subfigure}[t]{0.243\linewidth}
    \includegraphics[width=\linewidth]{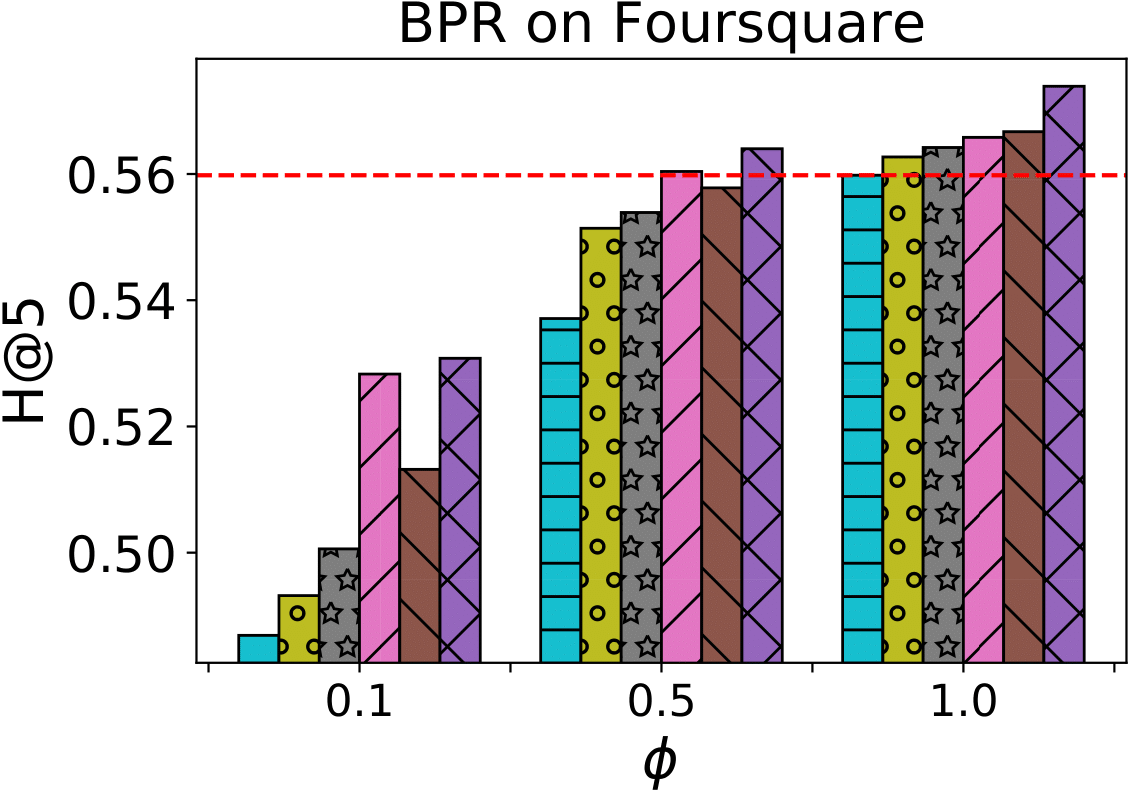}
\end{subfigure}
\begin{subfigure}[t]{0.243\linewidth}
    \includegraphics[width=\linewidth]{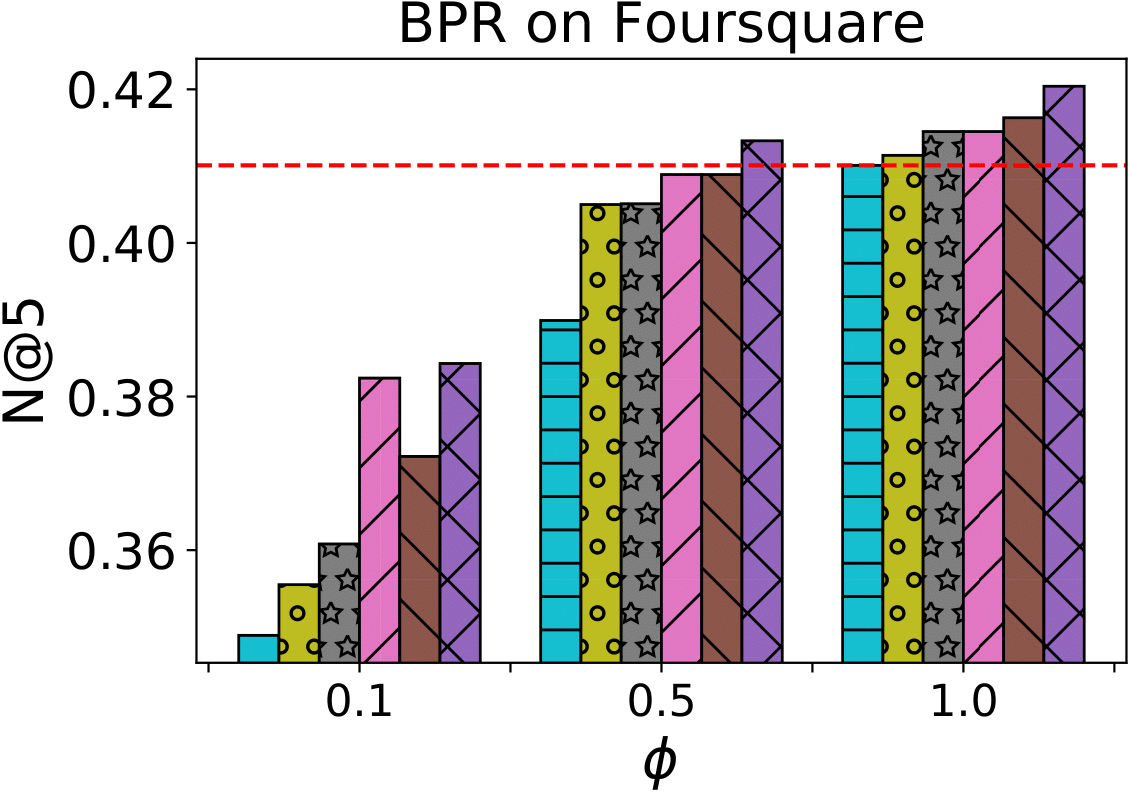}
\end{subfigure} 
\begin{subfigure}[t]{0.243\linewidth}
    \includegraphics[width=\linewidth]{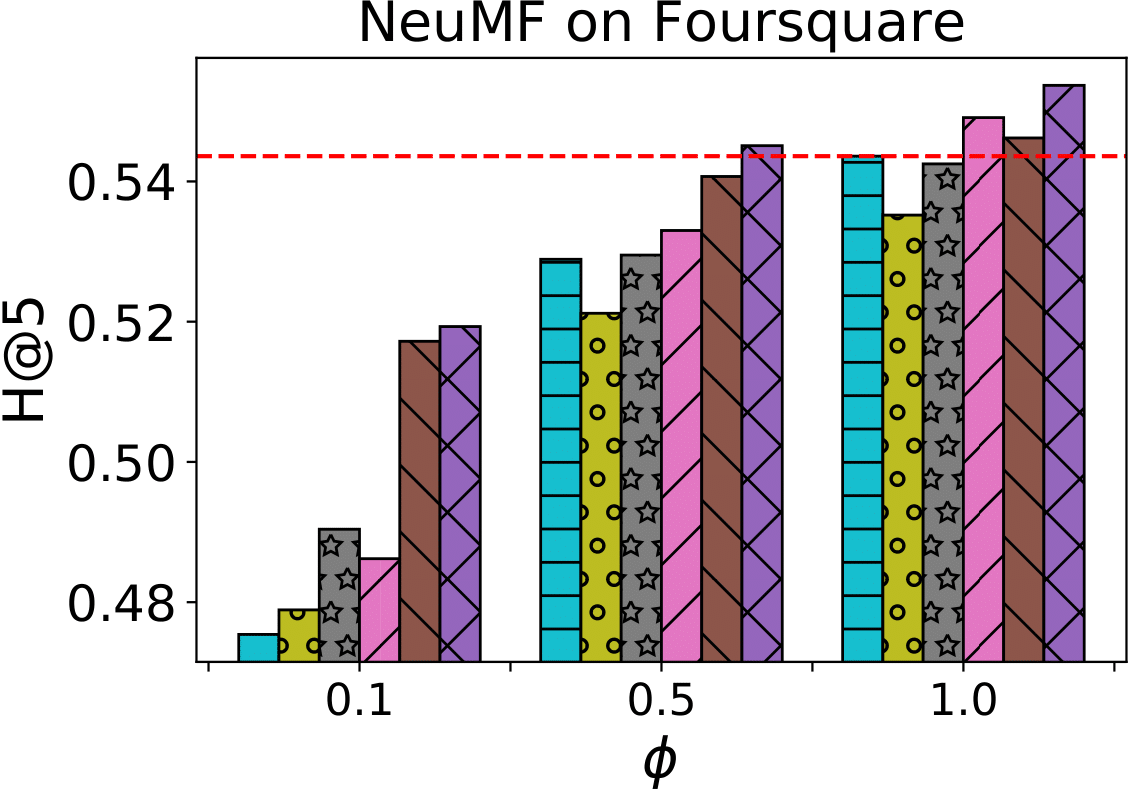}
\end{subfigure} 
\begin{subfigure}[t]{0.243\linewidth}
    \includegraphics[width=\linewidth]{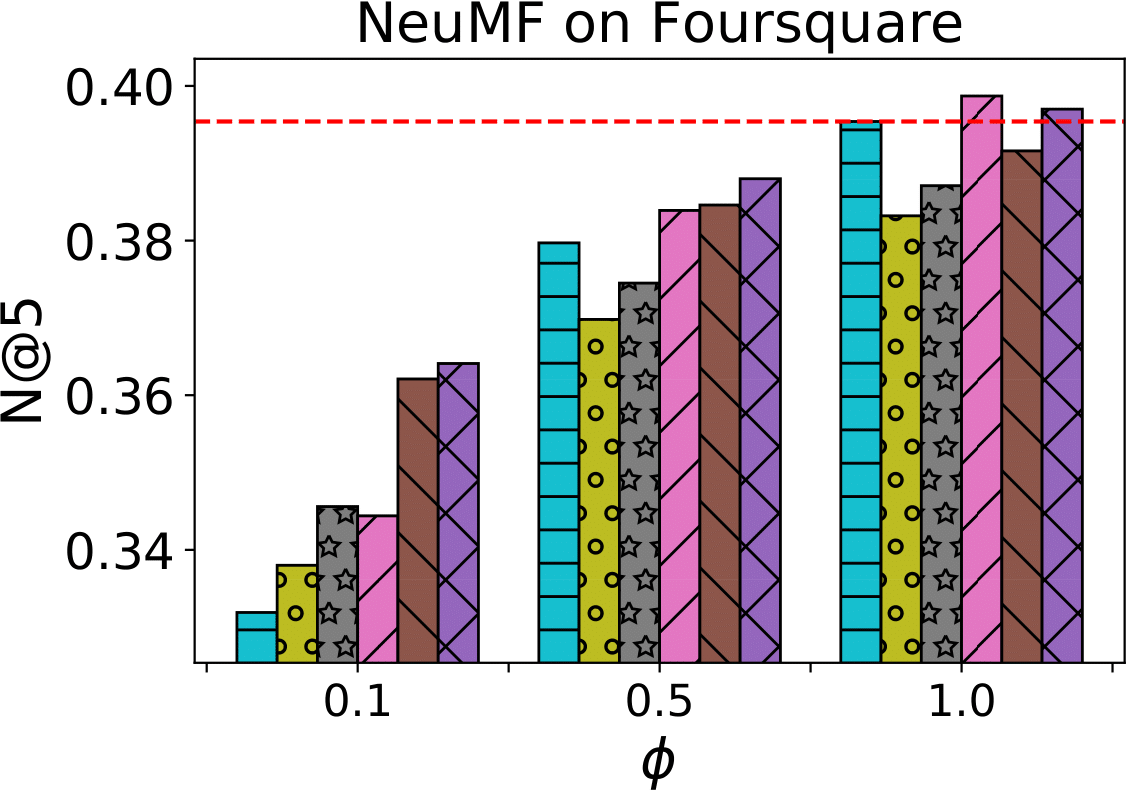}
\end{subfigure} 
\caption{Recommendation Performance on across three different student model sizes. (Red dotted line: Teacher)}
\label{fig:sizes}
\end{figure*}

\vspace{-0.2cm}
\subsection{Performance Comparison}
Table \ref{tab:main} shows top-$N$ recommendation accuracy of different methods in terms of various ranking metrics.
In summary, DE-RRD shows the significant improvement compared to the state-of-the-art KD methods on two base models that have different architectures and optimization strategies.
Also, DE-RRD consistently outperforms the existing methods on three different sizes of the student model in Figure \ref{fig:sizes}.
We analyze the results from various perspectives.

We first observe that the two methods of the proposed framework (i.e., DE, RRD) improve the performance of the student model. 
DE directly distills the teacher's latent knowledge that includes detailed information on users, items, and the relationships among them.
This enables the student to be more effectively trained than finding such information from scratch with a limited capacity.
RRD distills the knowledge from the teacher's predictions based on the relaxed ranking approach which makes the student to effectively maintain the ranking orders of interesting items predicted by the teacher. 
Unlike the existing methods (i.e., RD, CD), it directly handles the ranking violations among the sampled items, which can lead to better ranking performance.

Also, we observe that RRD achieves large performance gain particularly in NeuMF ($\phi=0.1$).
One possible reason is that NeuMF is trained with the point-wise loss function (i.e., binary cross-entropy) which considers only one item at a time.
In general, it is known that the approaches considering the preference orders between items (e.g., pair-wise, list-wise) can achieve better ranking performance than the point-wise approach \cite{NCR}.
RRD enables the model to capture the ranking orders among the unobserved items, so that it can lead to the large performance gain.
Interestingly, we observe that the prediction-based KD methods (i.e., RD, CD, RRD) can have an adverse effect when the model size is large (NeuMF with $\phi = 0.5, 1.0$ in Figure \ref{fig:sizes}).
We conjecture that this is because when a model has sufficient capacity to achieve comparable performance to the teacher, enforcing it to exactly mimic the teacher's prediction results can act as a strong constraint that rather hinders~its~learning.

In addition, we observe that DE-RRD achieves the best performance among all the methods in general.
DE-RRD enables the student to learn both from the teacher's prediction and from the latent knowledge that provides the bases for such predictions.
Interestingly, DE-RRD also shows a large performance gain when the student model has the identical structure to the teacher model (i.e., self-distillation with $\phi=1.0$ in Figure \ref{fig:sizes}).
This result shows that it can be also used to maximize the performance of the existing recommender.

Lastly, we provide the result of the online inference efficiency test in Table \ref{table:latency}.
All inferences are made using PyTorch with CUDA from Tesla P40 GPU and Xeon on Gold 6148 CPU.
The student model trained with DE-RRD achieves comparable performance with only 10-50\% of learning parameters compared to the teacher.
The smaller model requires less computations and memory costs, so it can achieve lower latency.
In particular, deep recommender (i.e., NeuMF) which has a large number of learning parameters and complex structures takes more benefits from the smaller model size.
On real-time RS application that has larger numbers of users (and items) and has a more complex model structure, DE-RRD can lead to a larger improvement in online inference efficiency.

\begin{table}[t]
\small
\RowStretch{0.3}
  \caption{Model compactness and online inference efficiency. Time (seconds) indicates the wall time used for generating recommendation list for every user. 
  H@5 Ratio denotes the ratio of H@5 from DE-RRD over that from Teacher.}
  \begin{tabular}{cccccc}
  \toprule
    Dataset & Base Model & $\phi$ & Time (s) & \#Params.& H@5 Ratio\\
   \midrule
   &\multirow{3}*{BPR}&1.0&59.27s&6.08M& 1.03\\
   &&0.5&57.53s&3.04M& 1.01\\
   \multirow{4}*{\rotatebox[origin=c]{0}{CiteULike}}&&0.1&55.39s&0.61M& 0.94\\
   \cmidrule{2-6}
   &\multirow{3}*{NeuMF}&1.0&79.27s&15.33M&1.01\\
   &&0.5&68.37s&7.63M& 1.01\\
   &&0.1&58.27s&1.52M& 0.99\\
   \midrule
&\multirow{3}*{BPR}&1.0&257.28s&9.61M& 1.03\\
   &&0.5&249.19s&4.81M& 1.01\\
   \multirow{4}*{\rotatebox[origin=c]{0}{Foursquare}}&&0.1&244.23s&0.96M& 0.95\\
   \cmidrule{2-6}
   &\multirow{3}*{NeuMF}&1.0&342.84s&24.16M& 1.02\\
   &&0.5&297.34s&12.05M&1.01\\
   &&0.1&255.24s&2.40M&0.95\\
   \bottomrule
  \end{tabular}
  \label{table:latency}
\end{table}

\subsection{Design Choice Analysis}
We provide both quantitative and qualitative analyses on the proposed methods and alternative design choices (i.e., ablations) to verify the superiority of our design choice.
The performance comparisons with the ablations are summarized in Table \ref{table:ablations}.

For DE, we consider three ablations: (a) Attention (b) One expert (large) (c) One expert (small).
As discussed in Section 4.1.2, instead of the selection strategy, attention mechanism can be adopted.
We also compare the performance of one large expert\footnote{
We make one large expert by adopting the average pooling.} 
and one small expert.
Note that DE, attention, and one expert (large) has the exact same number of learning parameters for experts.
We observe that the increased numbers of learning parameters do not necessarily contribute to performance improvement ((a) vs. (c) in BPR).

\begin{figure}[t]
\vspace{-0.3cm}
\begin{subfigure}[t]{0.5\linewidth}
    \includegraphics[width=\linewidth]{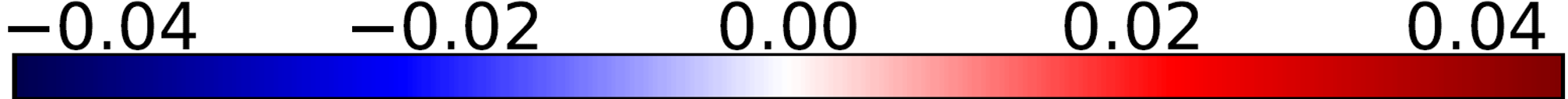}
\end{subfigure} 
\\
\begin{subfigure}[t]{0.33\linewidth}
    \includegraphics[width=\linewidth]{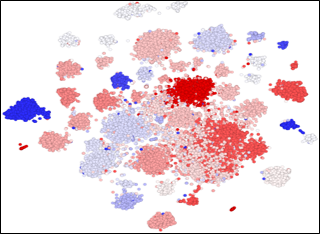}
    \caption*{Attention}
\end{subfigure} 
\hspace*{-0.07in}
\begin{subfigure}[t]{0.33\linewidth}
    \includegraphics[width=\linewidth]{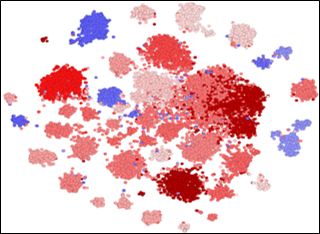}
    \caption*{One expert (large)}
\end{subfigure}
\hspace*{-0.07in}
\begin{subfigure}[t]{0.33\linewidth}
    \includegraphics[width=\linewidth]{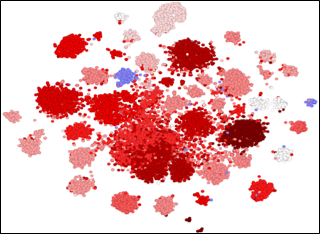}
    \caption*{DE (selection)}
\end{subfigure} 
\caption{Performance (N@20) gain map (BPR with $\phi=0.1$ on Foursquare).}
\label{fig:pgain}

\begin{subfigure}[t]{0.33\linewidth}
    \includegraphics[width=\linewidth]{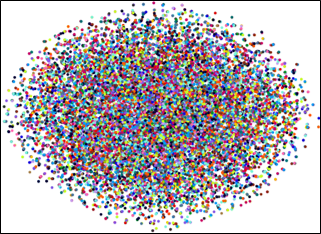}
    \caption{Epoch 0}
\end{subfigure} 
\hspace*{-0.07in}
\begin{subfigure}[t]{0.33\linewidth}
    \includegraphics[width=\linewidth]{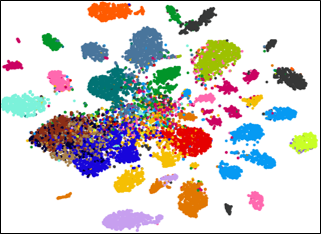}
    \caption{Epoch 20}
\end{subfigure}
\hspace*{-0.07in}
\begin{subfigure}[t]{0.33\linewidth}
    \includegraphics[width=\linewidth]{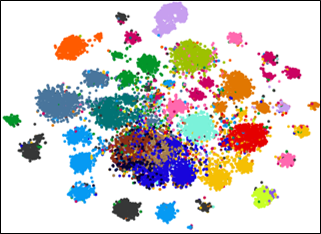}
    \caption{Train Done}
\end{subfigure} 
\caption{Expert selection map of DE. Each color corresponds to an expert (BPR with $\phi=0.1$ on Foursquare).}
\label{fig:selection_map}
\end{figure}

We also observe that the selection shows the best performance among all the ablations.
To further investigate this result, we conduct qualitative analysis on user representation spaces induced by each design choice.
Specifically, we first perform clustering\footnote{We use $k$-Means clustering in Scikit-learn. $k$ is set to 20.} on user representation space from the teacher model to find user groups that have strong correlations (or similar preferences).
Then, we visualize the average performance gain (per group) map in Figure \ref{fig:pgain}.
We observe that distilling the knowledge by the attention, one large expert can cause performance decreases in many user groups (blue clusters), whereas the selection improves the performance in more numbers of user groups (red clusters).
In the ablations (a)-(c), the experts are trained to minimize the overall reconstruction errors on all the diverse entities.
This makes the information of weakly correlated entities to be mixed together and further hinders discovering the preference of a particular user group.
Unlike the ablations, DE clearly distinguishes the knowledge that each expert distills, and makes each expert
to be trained to distill only the knowledge of strongly correlated entities.
So, it can alleviate such problem.
The expert selection map of DE is visualized in Figure \ref{fig:selection_map}.
We can observe that each expert gets gradually specialized in certain user groups that share similar preferences during the training.
\begin{table}[t!]
\vspace*{-0.2cm}
 \caption{Performance comparison for alternative design choices on Foursquare ($\phi=0.1$).}
\small
\RowStretch{0.15}
\setlength\tabcolsep{2pt}
\resizebox{\columnwidth}{!}{%

  \begin{tabular}{clcccc}
   
  \toprule
    Base Model & Design choices & H@5 & N@5 & H@10 & N@10\\
  \midrule
  &DE&\textbf{0.5283}&\textbf{0.3824}&\textbf{0.6810}&\textbf{0.4316}\\
  &(a) Attention&0.5019&0.3625&0.6575&0.4131\\
  \multirow{5}*{BPR}&(b) One expert (large)&0.5151&0.3716&0.6733&0.4230\\
  &(c) One expert (small)&0.5136&0.3717&0.6683&0.4213\\
  \cmidrule{2-6}
  &RRD&\textbf{0.5132}&\textbf{0.3722}&\textbf{0.6616}&\textbf{0.4202}\\
  &(d) Full ranking &0.4983&0.3595&0.6474&0.4080\\
  &(e) \textit{Interesting} ranking & 0.4814&0.3479&0.6416&0.3999\\
  \midrule
  &DE&\textbf{0.4862}&\textbf{0.3444}&\textbf{0.6413}&\textbf{0.3938}\\
  &(a) Attention&0.4770&0.3364&0.6364&0.3903\\
  \multirow{5}*{NeuMF}&(b) One expert (large)&0.4741&0.3367&0.6341&0.3885\\
  &(c) One expert (small)&0.4740&0.3339&0.6316&0.3860\\
  \cmidrule{2-6}
  &RRD&\textbf{0.5172}&\textbf{0.3621}&\textbf{0.6739}&\textbf{0.4132}\\
  &(d) Full ranking &0.4799&0.3457&0.6324&0.3949\\
  &(e) \textit{Interesting} ranking & 0.4641&0.3294&0.6228&0.3809\\
  \bottomrule
  \end{tabular}%
  }
  \label{table:ablations}
\end{table}

For RRD, we consider two ablations: (d) and (e).
The ablations are intended to show the effects of the proposed relaxed ranking.
Concretely, we apply the list-wise loss (i.e., no relaxation) on all the sampled items (interesting and uninteresting items) for (d), on the top-ranked items (interesting items) for (e).
Note that all the methods use the same number of items for distillation.
We observe that merely adopting the list-wise loss has adverse effects on the ranking performance.
First, (d) learns to match the full ranking order among all the sampled items.
Learning the detailed order among the uninteresting items is not necessarily helpful to improve the ranking performance, and may further interfere with focusing on the interesting items.
Also, (e), which only considers the interesting items, shows even worse performance than Student.
The list-wise loss does not take into account the absolute ranking positions of the items; a ranking order can be satisfied regardless of the items' absolute positions.
Since (e) does not consider the relative orders between the interesting items and the uninteresting items, it may push such interesting items far from the top of the ranking list.
Unlike the ablations, RRD adopts the relaxed ranking approach, which enables the student to better focus on the interesting items while considering the relative orders with the uninteresting items.

\subsection{Hyperparameter Analysis}
\begin{figure}[t]
\begin{subfigure}[t]{0.25\linewidth}
    \includegraphics[width=\linewidth, height=0.9\linewidth]{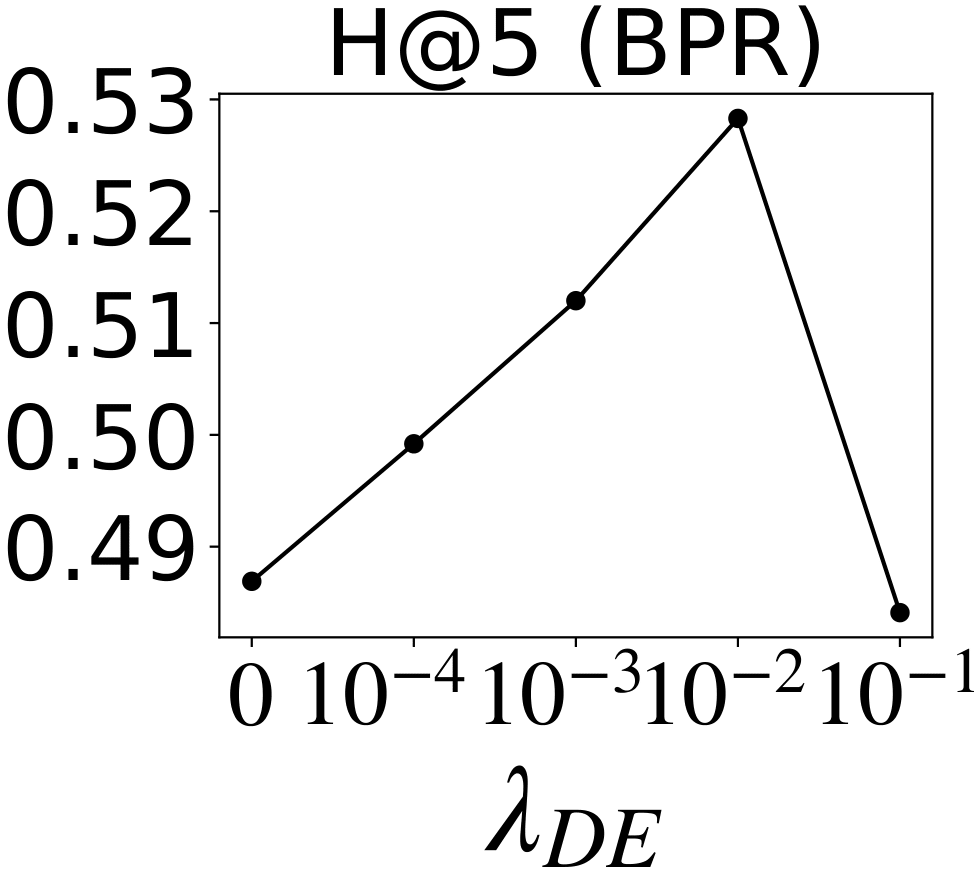}
\end{subfigure}
\hspace*{-0.05in}
\begin{subfigure}[t]{0.25\linewidth}
    \includegraphics[width=\linewidth, height=0.9\linewidth]{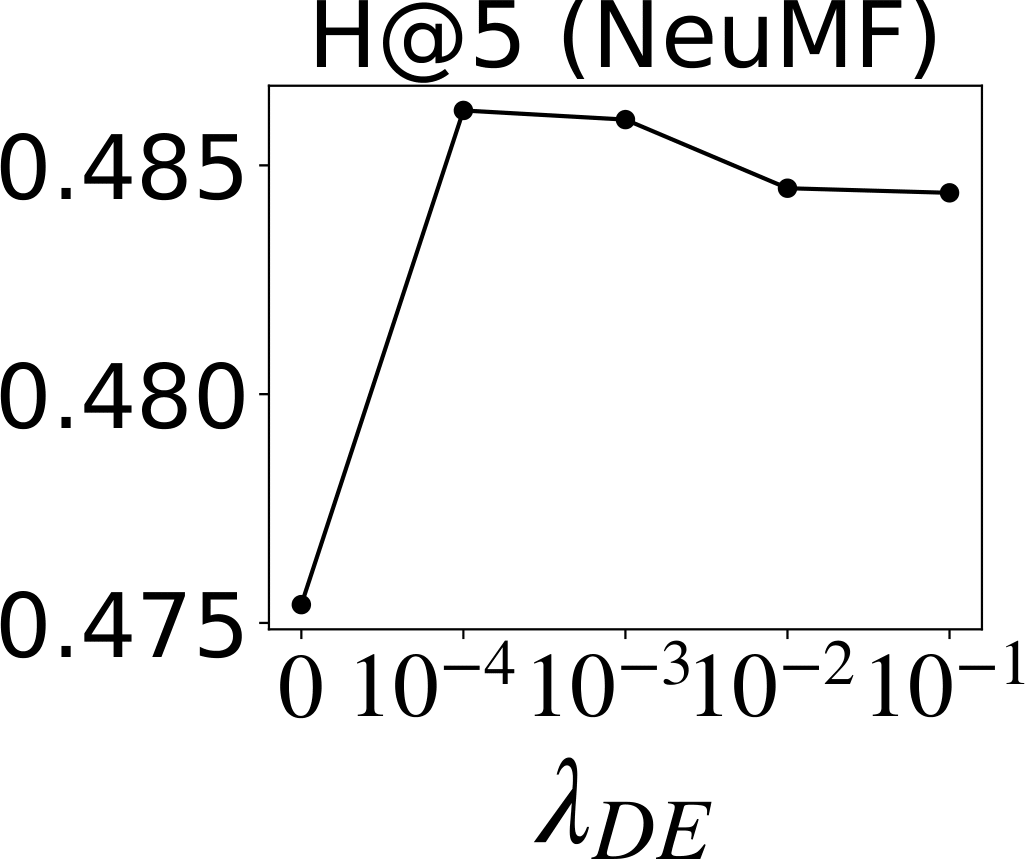}
\end{subfigure} 
\hspace*{-0.05in}
\begin{subfigure}[t]{0.25\linewidth}
    \includegraphics[width=0.98\linewidth, height=0.9\linewidth]{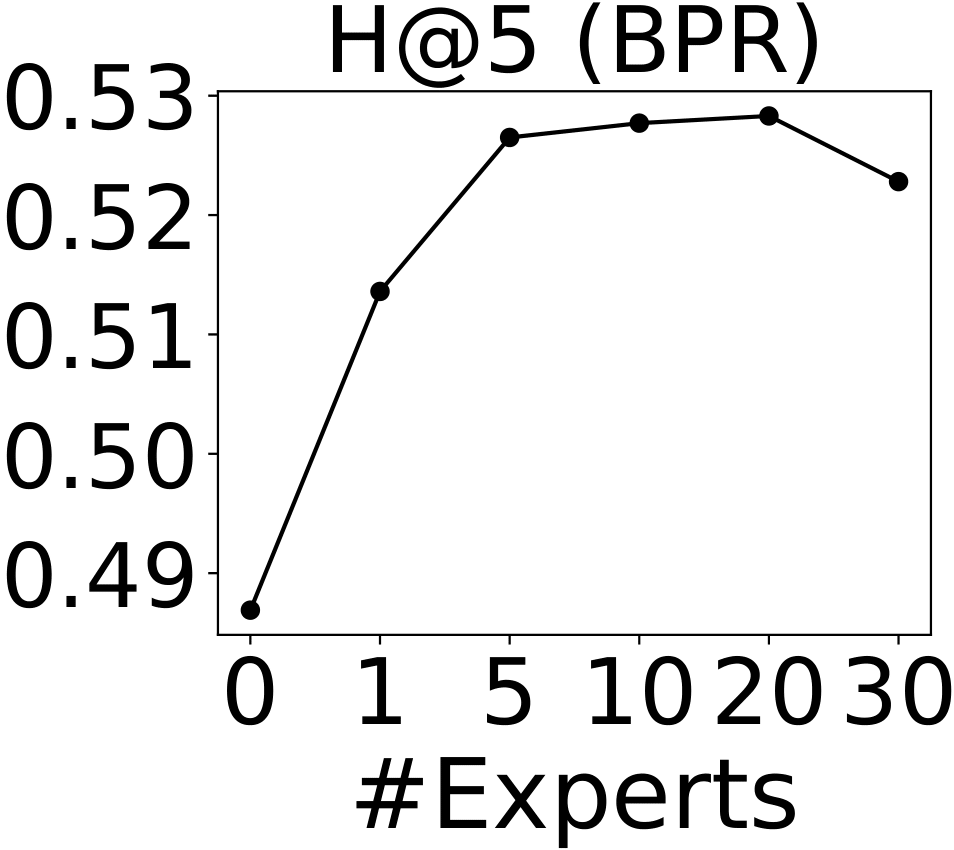}
\end{subfigure} 
\hspace*{-0.07in}
\begin{subfigure}[t]{0.25\linewidth}
    \includegraphics[width=\linewidth, height=0.9\linewidth]{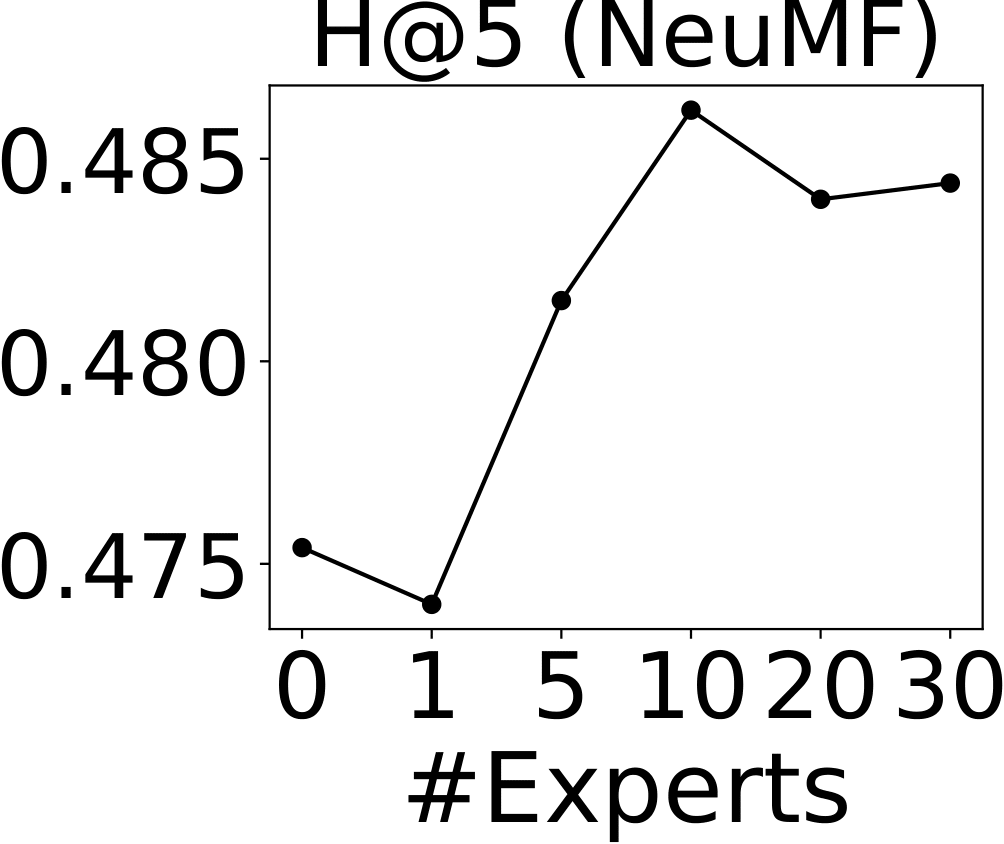}
\end{subfigure} 
\caption*{(a) Effects of $\lambda_{DE}$ and the number of experts\quad}
\vspace*{0.05in}
\begin{subfigure}[t]{0.25\linewidth}
    \includegraphics[width=\linewidth, height=0.9\linewidth]{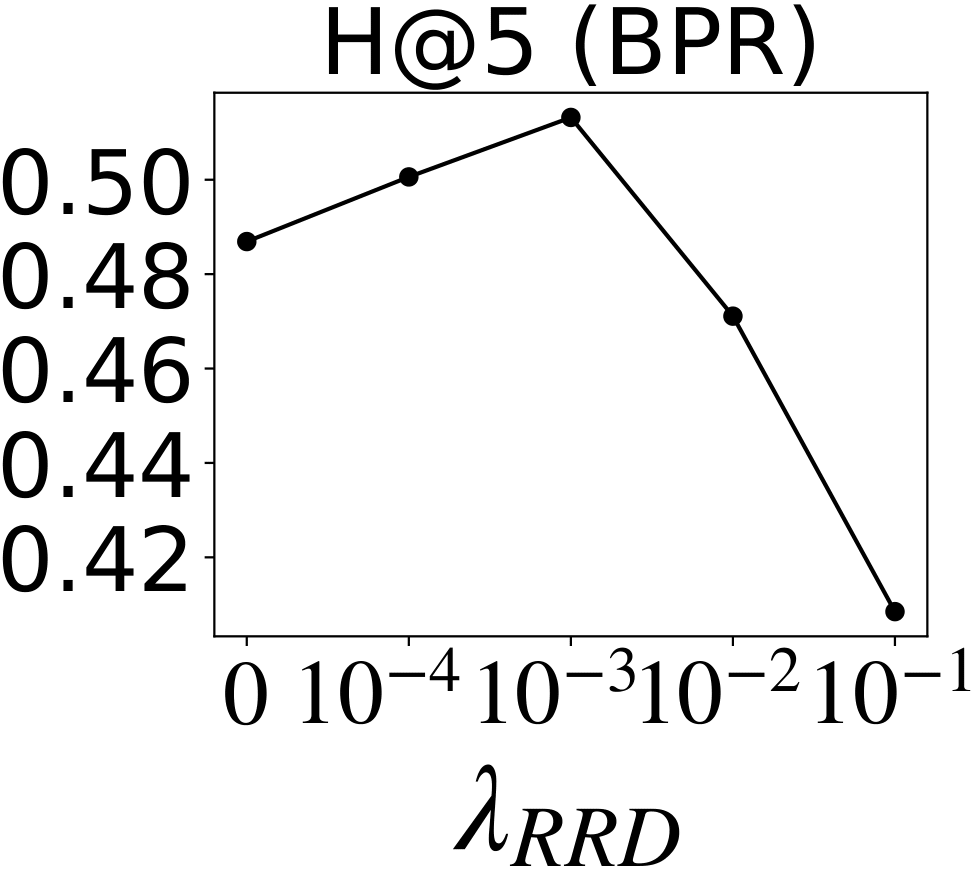}
\end{subfigure}
\hspace*{-0.05in}
\begin{subfigure}[t]{0.25\linewidth}
    \includegraphics[width=\linewidth, height=0.9\linewidth]{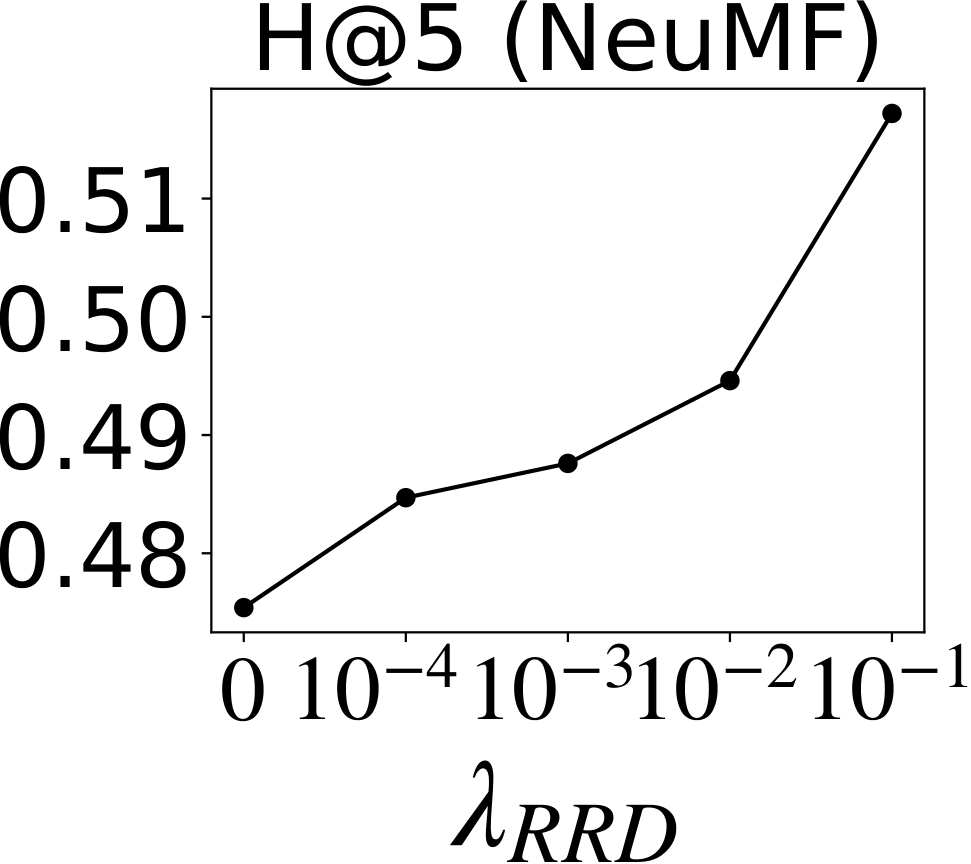}
\end{subfigure} 
\hspace*{-0.05in}
\begin{subfigure}[t]{0.25\linewidth}
    \includegraphics[width=0.98\linewidth, height=0.9\linewidth]{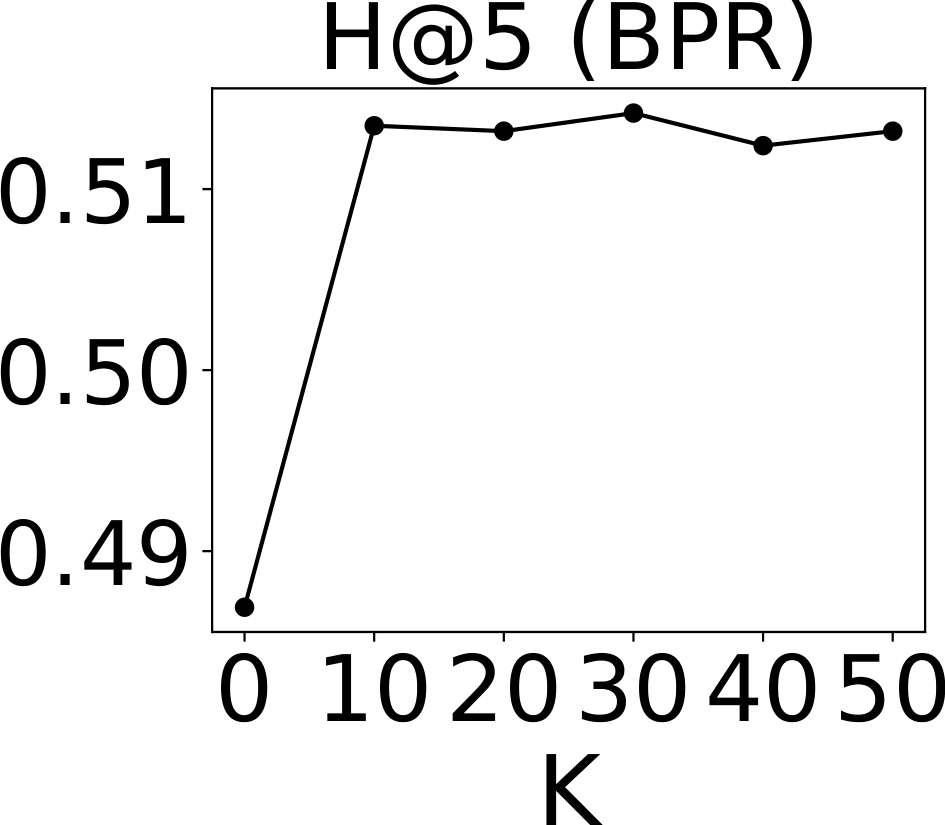}
\end{subfigure} 
\hspace*{-0.07in}
\begin{subfigure}[t]{0.25\linewidth}
    \includegraphics[width=\linewidth, height=0.9\linewidth]{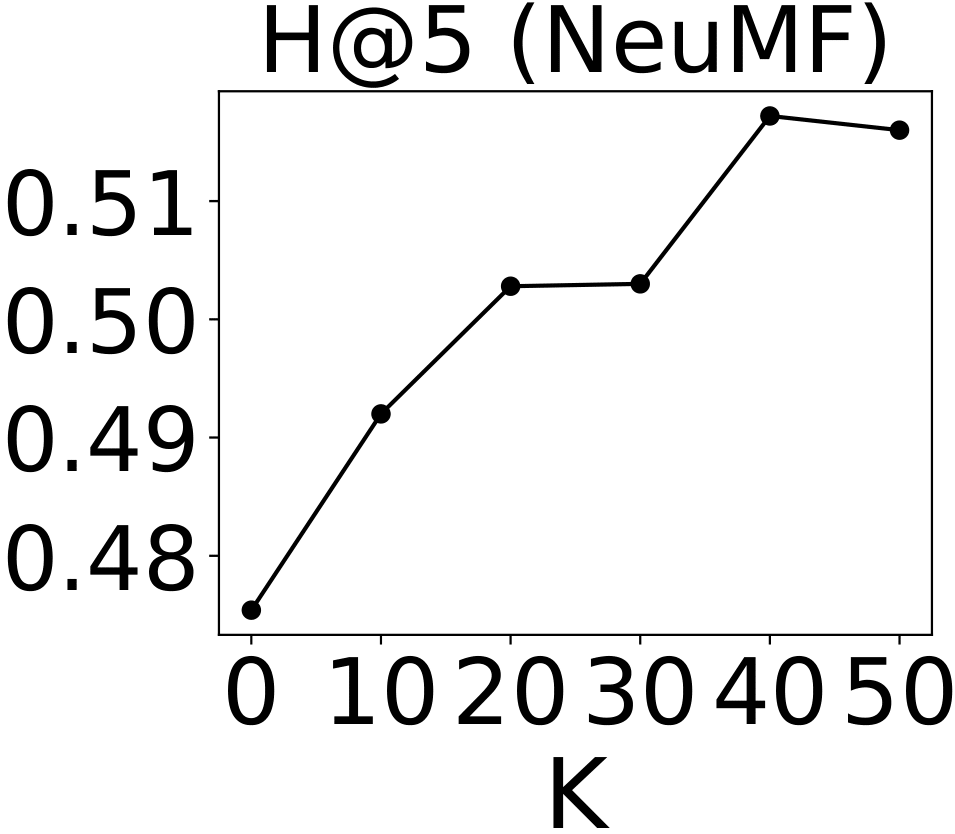}
\end{subfigure} 
\caption*{(b) Effects of $\lambda_{RRD}$ and $K$\quad}
\caption{Effects of the hyperparameters. (a) DE (b) RRD.}
\label{fig:hp}
\end{figure}

We provide analyses to offer guidance of hyperparameter selection of DE-RRD.
For the sake of space, we report the results on Foursquare dataset with $\phi=0.1$.
We observe similar tendencies on CiteULike dataset.
For DE, we show the effects of two hyperparameters: $\lambda_{DE}$ that controls the importance of DE and the number of experts in Figure \ref{fig:hp}a.
For RRD, we show the effects of two hyperparameters: $\lambda_{RRD}$ that controls the importance of RRD and the number of interesting items ($K$) in Figure \ref{fig:hp}b. 
In our experiment, the number of uninteresting items is set to the same with $K$.
Note that for all graphs value ‘0’ corresponds to Student (i.e., no distillation).

Because the types of loss function of the proposed methods are different from that of the base models, it is important to properly balance the losses by using $\lambda$. 
For DE, the best performance is achieved when the magnitude of DE loss is approximately 20\% (BPR), 2-5\% (NeuMF) compared to that of the base model’s loss. 
For RRD, the best performance is achieved when the magnitude of RRD loss is approximately 7-10\% (BPR), 1000\% (NeuMF) compared to that of the base model’s loss. 
For the number of experts and $K$, the best performance is achieved near 10-20 and 30-40, respectively. 
Lastly, we show the effects of combinations of $\lambda_{DE}$ and $\lambda_{RRD}$ in DE-RRD framework in Table \ref{tab:hp}.
Generally, the best performance of DE-RRD is observed in the ranges where each method (i.e., DE, RRD) achieves the best performance.

\section{Conclusion}
\label{sec:conclusion}

\begin{table}[t]
\vspace{-0.5cm}
\caption{Effects of $\lambda_{DE}$ and $\lambda_{RRD}$ in DE-RRD framework.}
\resizebox{1.01\columnwidth}{!}{%
\RowStretch{0.3}
\setlength\tabcolsep{2.5pt}
\begin{tabular}{|c|c|cccc|cccc|}
\hline
\multicolumn{2}{|c|}{} & \multicolumn{4}{c|}{\textbf{BPR}} & \multicolumn{4}{c|}{\textbf{NeuMF}} \\ \cline{3-10} 
\multicolumn{2}{|c|}{\multirow{-3}{*}{\begin{tabular}[c]{@{}c@{}}Foursquare\\ (H@5)\\\quad\end{tabular}}} & \multicolumn{4}{c|}{{\color[HTML]{000000} $\lambda_{RRD}$}} & \multicolumn{4}{c|}{$\lambda_{RRD}$} \\ \cline{3-10} 
 \multicolumn{2}{|c|}{}& $10^{-4}$ & $10^{-3}$ & $10^{-2}$ & $10^{-1}$ & $10^{-4}$ & $10^{-3}$ & $10^{-2}$ & $10^{-1}$ \\ \hline
\multicolumn{1}{|c|}{} & $10^{-4}$ & 0.5081 & 0.5201 & 0.4590 & 0.3901 & 0.4774 & 0.4896 & 0.5014 & \textbf{0.5193} \\
\multicolumn{1}{|c|}{\multirow{-2}{*}{\rotatebox[origin=c]{90}{$\lambda_{DE}$}}} & $10^{-3}$ & 0.5186 & 0.5276 & 0.4688 & 0.3906 & 0.4774 & 0.4858 & 0.4942 & 0.5112 \\
\multicolumn{1}{|c|}{} & $10^{-2}$ & 0.5261 & \textbf{0.5308} & 0.4791 & 0.3977 & 0.4846 & 0.4868 & 0.4892 & 0.5110 \\
\multicolumn{1}{|c|}{} & $10^{-1}$ & 0.5269 & \textbf{0.5308} & 0.4928 & 0.4154 & 0.4848 & 0.4881 & 0.4908 & 0.5055 \\ \hline
\end{tabular}%
}
\label{tab:hp}
\end{table}
This paper proposes a novel knowledge distillation framework for recommender system, DE-RRD, that enables the student model to learn both from the teacher's predictions and from the latent knowledge stored in a teacher model.
To this end, we propose two novel methods:
1) DE that directly distills latent knowledge from the representation space of the teacher.
DE adopts the experts and the expert selection strategy to effectively distill the vast CF knowledge to the student. 
2) RRD that distills knowledge revealed from teacher's predictions with direct considerations of ranking orders among items.
RRD adopts the relaxed ranking approach to better focus on the interesting items.
Extensive experiment results demonstrate that DE-RRD significantly outperforms the state-of-the-art competitors.

\noindent
\textbf{Acknowledgements:}
This research was supported by 
the NRF grant funded by the MSIT: (No.~2017M3C4A7063570), 
the IITP grant funded by the MSIT: (No.~2018-0-00584), 
the IITP grant funded by the MSIP (No.~2019-0-01906, Artificial Intelligence Graduate School Program (POSTECH)) and
the MSIT under the ICT Creative Consilience program (IITP-2020-2011-1-00783) supervised by the IITP.

\vspace{-0.2cm}

\bibliographystyle{ACM-Reference-Format}
\bibliography{acmart}

\end{document}